\documentclass[runningheads]{llncs}

\usepackage{eccv}

\usepackage{eccvabbrv}

\usepackage{graphicx}
\usepackage{booktabs}

\usepackage{amsmath}
\usepackage{amssymb}
\usepackage{gensymb}

\usepackage{mathtools}

\newcommand{\supp}{supplemental material\xspace}

\newcommand{\mypara}[1]{\vspace{5pt}\noindent{\bf{#1}}}
\newcommand{\ours}{OmniFID}
\newcommand{\oursFull}{Omnidirectional FID}

\usepackage{hyperref}

\usepackage{orcidlink}

\begin{document}

\title{Geometry Fidelity for Spherical Images} 

\author{Anders Christensen\textsuperscript{\textdagger}\inst{,2,3}\orcidlink{0009-0009-0038-5485} \and
Nooshin Mojab\inst{1}\orcidlink{0000-0002-2687-9012} \and
Khushman Patel\inst{1}\orcidlink{0009-0006-7945-4874}
\and \\
Karan Ahuja\inst{1,4}\orcidlink{0000-0003-2497-0530}
\and
Zeynep Akata\inst{3,5,6}\orcidlink{0000-0002-1432-7747}
\and
Ole Winther\inst{2,7,8}\orcidlink{0000-0002-1966-3205}
\and \\
Mar Gonzalez-Franco\inst{1}\orcidlink{0000-0001-6165-4495}
\and
Andrea Colaco\inst{1}\orcidlink{0009-0001-6661-2216}}

\authorrunning{A.~Christensen et al.}

\institute{Google, USA\\
\and
Technical University of Denmark, Denmark\\
\and
Helmholtz Munich, Germany\\
\and
Northwestern University, USA\\
\and
Technical University of Munich, Germany\\
\and
Munich Center of Machine Learning, Germany\\
\and
University of Copenhagen, Denmark\\
\and
Copenhagen University Hospital, Denmark\\
\email{andchri@dtu.dk}, \email{nooshinmojab@google.com}
}

\maketitle
\def\thefootnote{\textdagger}\footnotetext{Work done at Google, USA}

\begin{abstract} 

Spherical or omni-directional images offer an immersive visual format appealing to a wide range of computer vision applications. However, geometric properties of spherical images pose a major challenge for models and metrics designed for ordinary 2D images.
Here, we show that direct application of Fréchet Inception Distance (FID) is insufficient for quantifying geometric fidelity in spherical images.
We introduce two quantitative metrics accounting for geometric constraints, namely \oursFull\ (\ours) and Discontinuity Score (DS). 
\ours\ is an extension of FID tailored to additionally capture field-of-view requirements of the spherical format by leveraging cubemap projections.
DS is a kernel-based seam alignment score of continuity across borders of 2D representations of spherical images. In experiments, \ours\ and DS quantify geometry fidelity issues that are undetected by FID.
\end{abstract}

\section{Introduction}
\label{sec:intro}
Spherical images, offering a full 360-degree horizontal and 180-degree vertical field of view, hold immense potential for a broad range of computer vision applications such as virtual reality, game design and immersive panoramic image viewing.
However, spherical images have geometric properties not exhibited by regular 2D images, and most datasets are not representative of this format of images. Consequently, most existing models are not directly applicable or optimized for spherical images. 
To reduce this challenge, we can project between a spherical 3D image and 2D representations of it. However, such projections present a series of trade-offs between conformity to the spherical image, and representing area of the sphere equally on the 2D plane. A great number of map projections have been developed to project spherical images to the plane. These projections have been designed to improve on specific properties, such as reducing distortions in the resulting viewpoint images \cite{chen2018recent, federico360video, hussain2021evaluation, equiangular}. For successful application, models applied on such representations must be aware of the inherent distortions in order to adhere to the geometric constraints of the 3D sphere. In this work we focus on equirectangular and cubemap projections as they are widely used in applications and graphics pipelines already \cite{zucker2018cube}.

\begin{figure}[t]
  \centering
  \begin{subfigure}{0.49\linewidth}
    \centering
    \includegraphics[height=2.86cm]{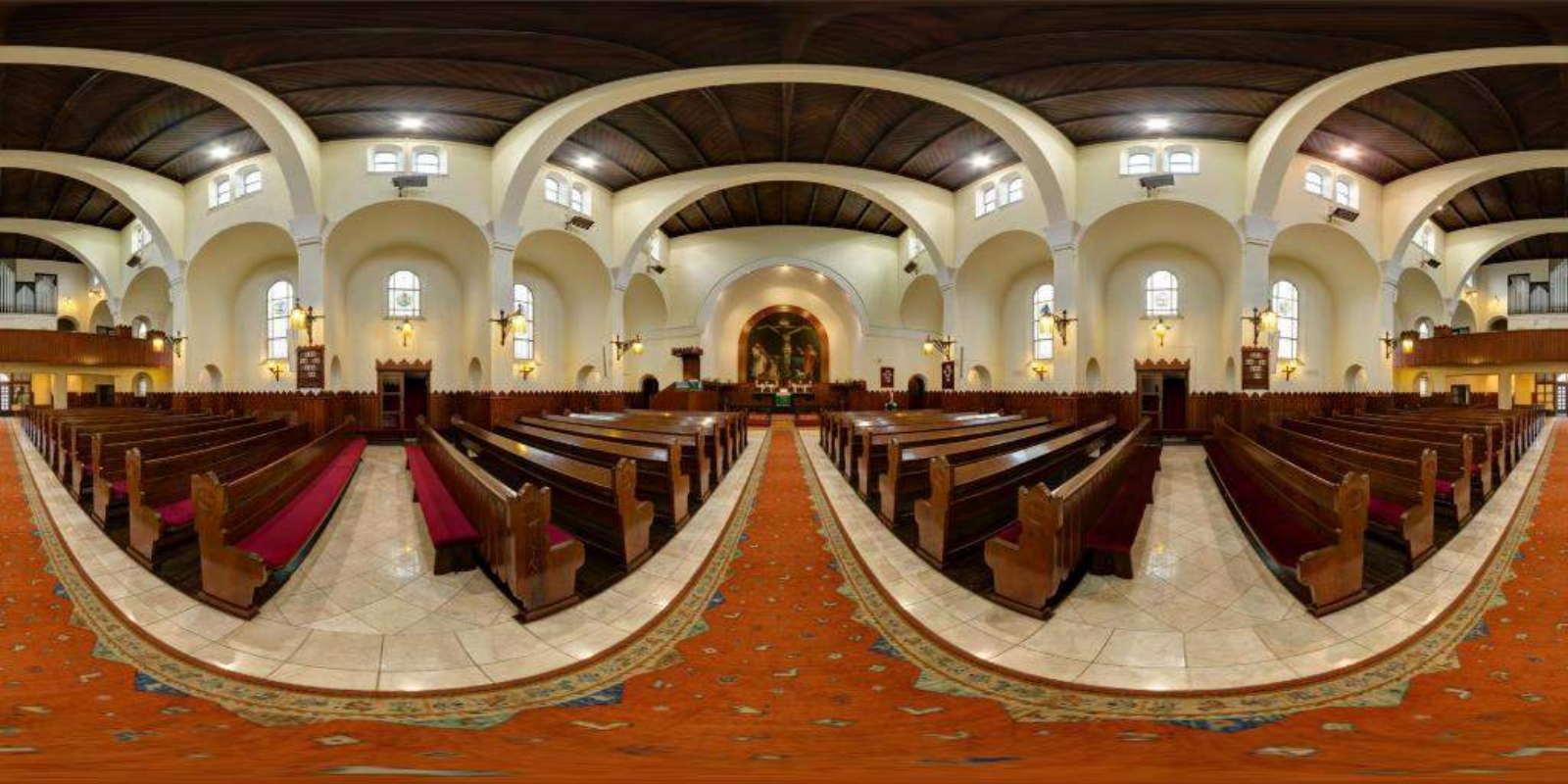}
    \label{fig:upscale-a}
  \end{subfigure}
  \hfill
  \begin{subfigure}{0.49\linewidth}
    \centering
    \includegraphics[height=2.86cm]{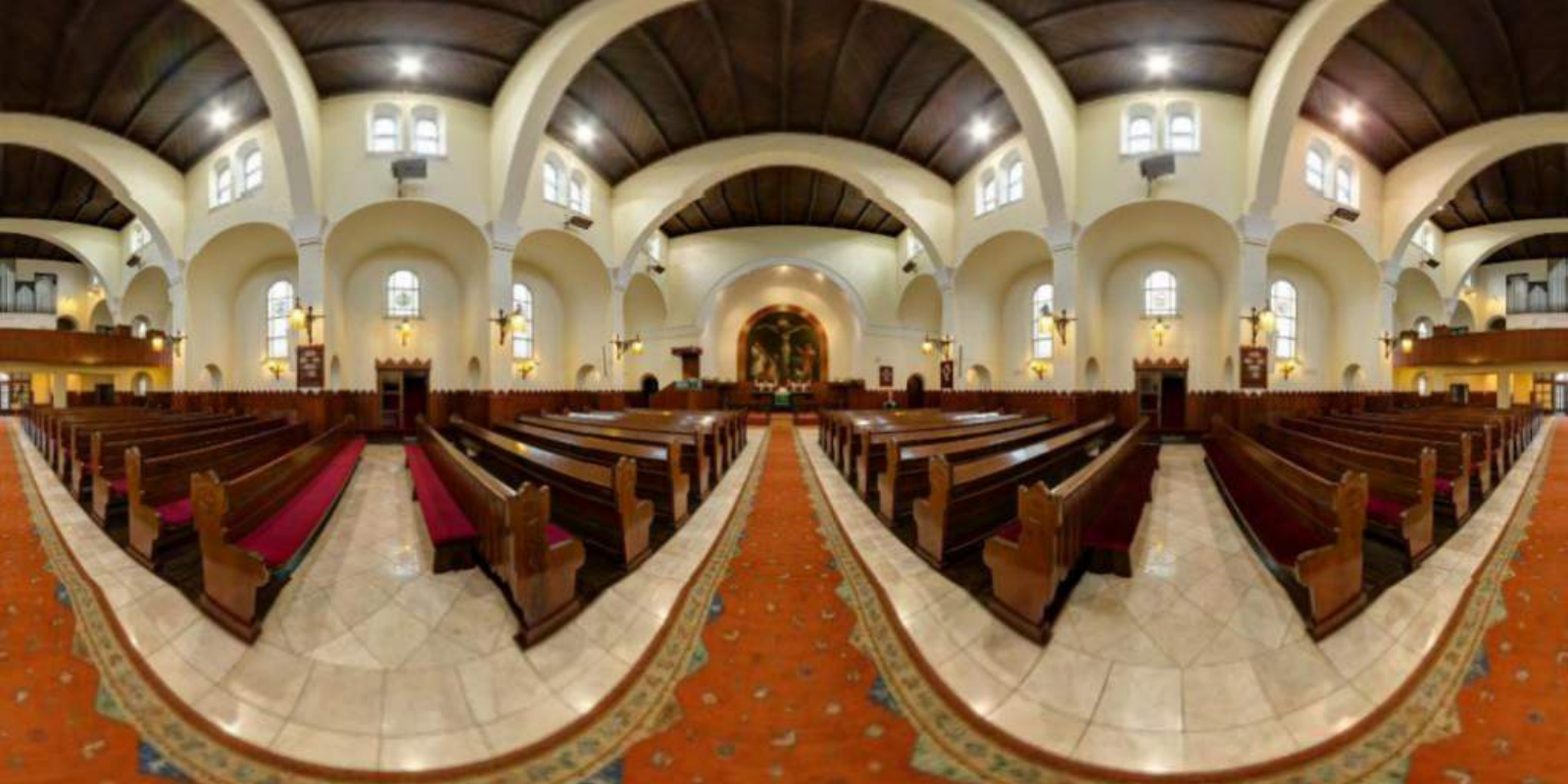}
    \label{fig:upscale-b}
  \end{subfigure}
  \begin{subfigure}{0.49\linewidth}
    \centering
    \includegraphics[height=2.8cm]{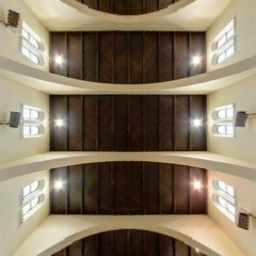}
    \includegraphics[height=2.8cm]{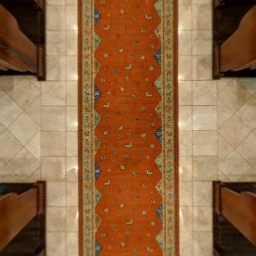}
    \label{fig:upscale-a}
  \end{subfigure}
  \hfill
  \begin{subfigure}{0.49\linewidth}
    \centering
    \includegraphics[height=2.8cm]{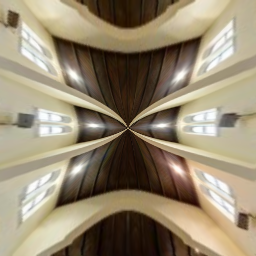}
    \includegraphics[height=2.8cm]{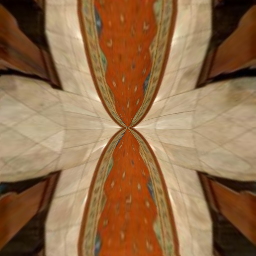}
    \label{fig:upscale-b}
  \end{subfigure}
  \caption{Visually, it is difficult to recognize field-of-view issues in the equirectangular format, but the problem is evident when rendered as a sphere and looking up/down. Top left: original spherical image with 180\degree\ vertical FOV represented as an equirectangular image. Top right: Resulting noisy equirectangular image, with a reduced vertical FOV of 140\degree. Bottom row: comparison of resulting views when looking upwards and downwards, respectively, in the two spherical images.}
  \label{fig:fov-issues-example}. 
\end{figure}

\begin{figure}[t]
  \centering
  \includegraphics[width=\textwidth]{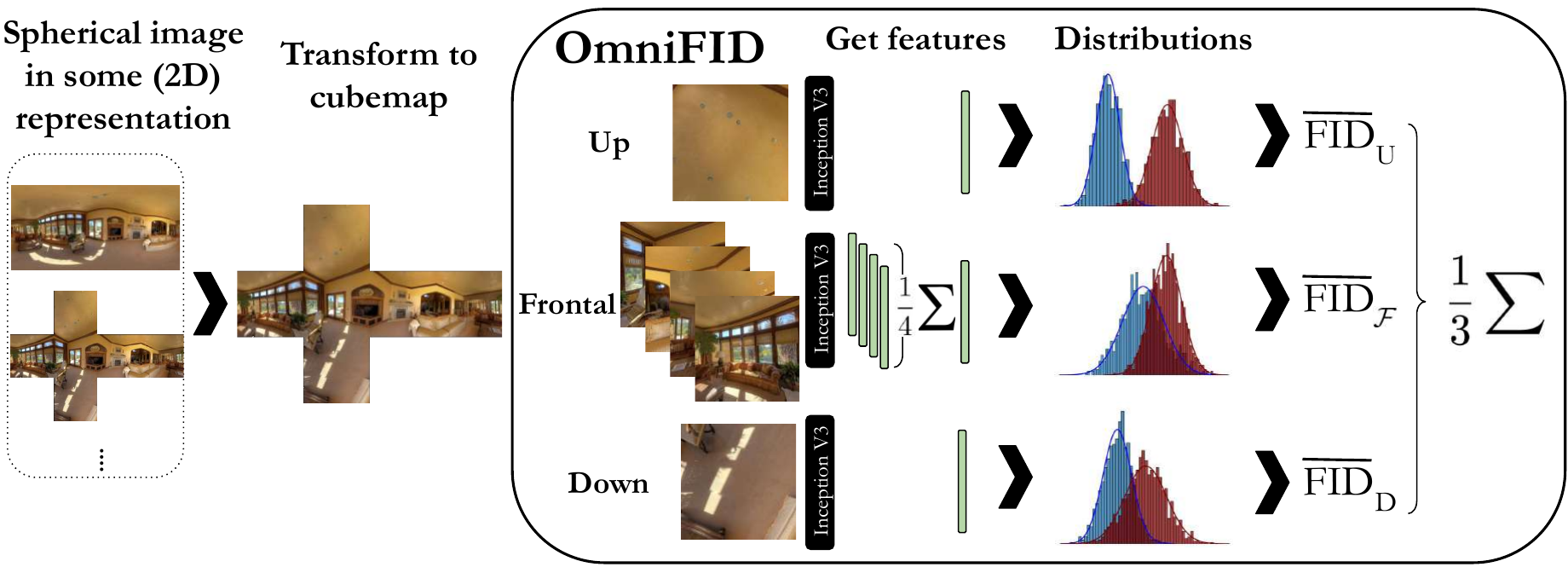}
  \caption{Visualisation of our proposed \oursFull. Utilizing cubemaps and using view-point dependent image features allows \ours\ to detect issues with the spherical geometry, such as insufficient field-of-view.
  }
  \label{fig:omnifid-vis}
\end{figure}

One of the key metrics to measure image fidelity for generative image models is FID \cite{Heusel2017GANsTB}. Having been reported in a range of works on generation of spherical images \cite{chen2022text2light, Lu2023AutoregressiveOO, wang2023360, akimoto2022diverse}, FID has also been established as the de-facto fidelity metric for spherical images. Our first contribution in this paper is to demonstrate that FID fails to capture distortions related to the unique geometric requirements of spherical images when applied on equirectangular projections. This is a severe limitation of the metric for spherical image applications. We show this by proposing a noise transformation of equirectangular images, effectively reducing the field of view with little distortion in its 2D equirectangular representation.  
Fundamentally, the FID metric relies on features extracted from the Inception V3 convolutional neural network trained on ordinary 2D linear perspective images \cite{szegedy2016rethinking}.
Hence, to increase compatibility of the underlying Inception network with spherical image data, we present an extension of FID, namely \oursFull. \ours\ utilizes cubemap representations as an alternative to the hitherto used equirectangular representations. Visually, cubemaps are often represented as a dice with its faces folded out (see e.g. \Cref{fig:omnifid-vis}). Unlike equirectangular images, cubemap views are the result of rectilinear projections, providing better conformity to the shapes in the actual spherical rendering. Through our experiments we showcase that \ours\ is able to capture reductions in field-of-view, a crucial aspect for a quality metric for spherical image generation, while maintaining other positive properties of FID, such as sensitivity to noise. 

Additionally, representing spherical images as 2D images introduces a requirement of continuity across image borders, otherwise resulting in visual seams. Unfortunately, 2D representations with better conformity increases the issue of border continuity \cite{dimitrijevic2016comparison}.
Therefore, assessing semantic alignment across borders of generated images is essential for a thorough evaluation of generated spherical images - even more so if the issue is amplified by using projections resulting in more seams, such as cubemaps. 
To this end, we additionally propose a simple kernel based algorithm for edge detection, which we refer to as Discontinuity Score (DS). DS measures the seam alignment across image borders.

In summary, we present two new metrics for geometry fidelity of generated spherical images, each capturing different aspects of geometric constraints unique to spherical images. \ours\ assesses the distortion related to field-of-view properties of spherical images utilizing cubemap representations, while DS assesses seam alignment across image borders. 

\begin{figure}
  \centering
  \includegraphics[width=0.98\textwidth]{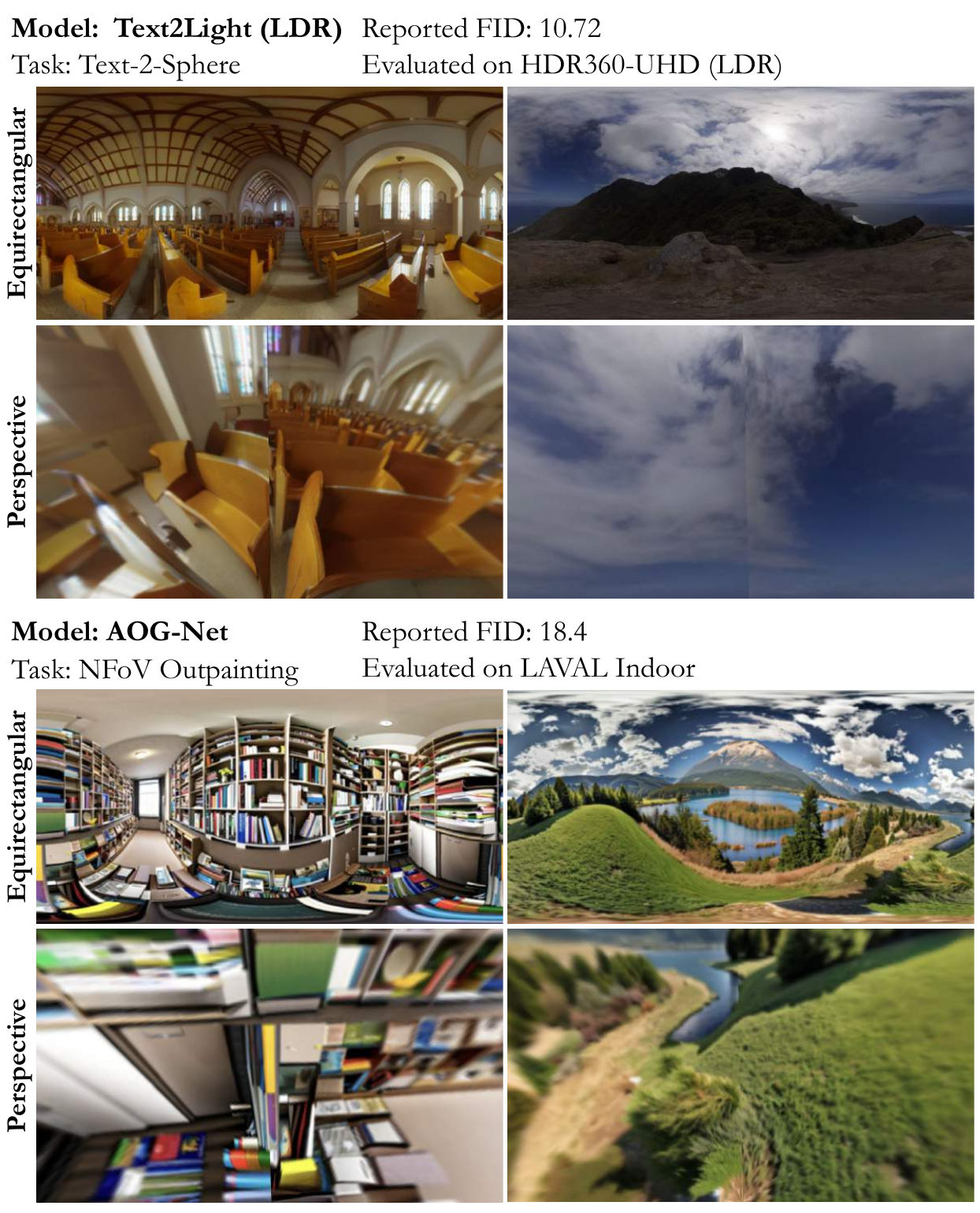}
  \caption{Although recent spherical image generation models (Text-2-Sphere and Image-2-Sphere) have begun achieving low FID scores, models are still struggling to produce images with full 180\degree\ vertical field-of-view and no seams. Above, we show equirectangular images from the models Text2Light \cite{chen2022text2light} and AOG-Net \cite{Lu2023AutoregressiveOO} (top row in each block), along with their reported FID score. These images are from their respective papers. Below each image we display a perspective view when looking backwards, showing the resulting stitching across image borders (and at the poles). We find that FID does not sufficiently capture geometry fidelity issues in the generated images, such as benches converging to a point at the poles, or inconsistencies across image borders.
  }
  \label{fig:sota-issues}
\end{figure}

\section{Related work}

\begin{figure}[t]
  \centering
  \includegraphics[height=4.5cm]{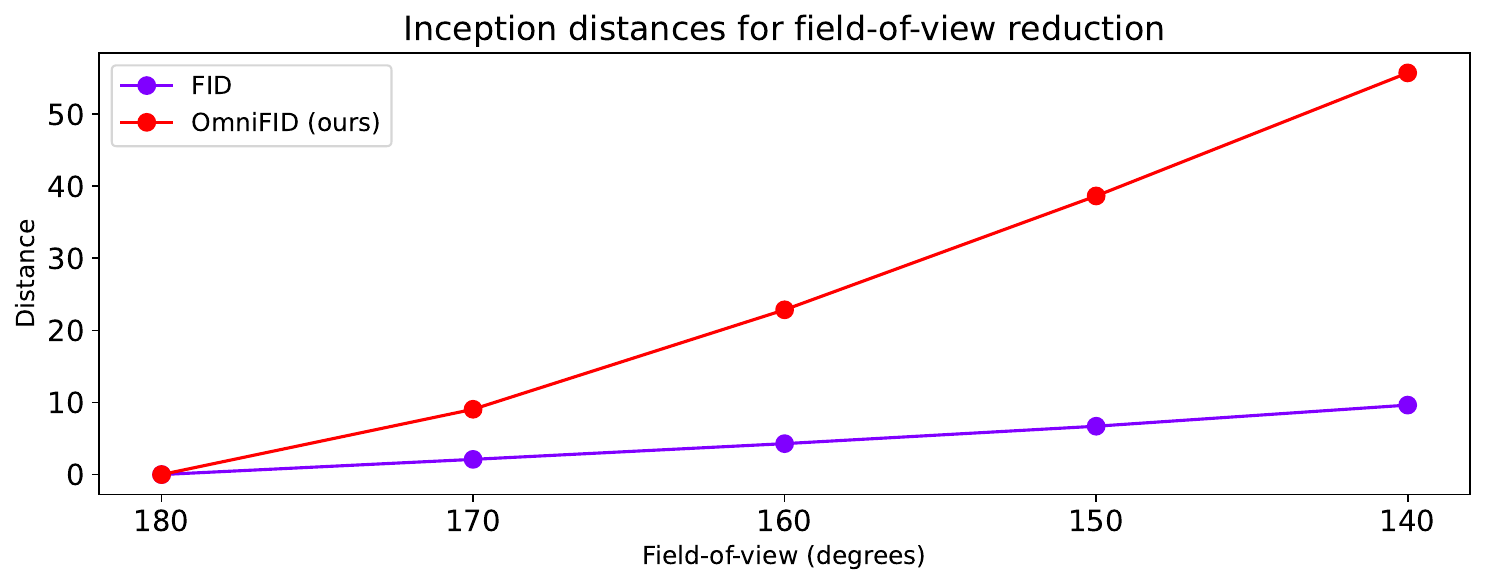}
  \caption{FID results compared to our modification, \ours, for detecting issues with field-of-view reductions on the 360-Indoor spherical image dataset \cite{Chou2019360IndoorTL}. FID increases negligibly, despite reducing the vertical field-of-view from 180\degree\ to 140\degree, while our proposed \ours\ captures the difference.  
  }
  \label{fig:exp-fov-noise}
\end{figure}

\mypara{Spherical Image Representation}
With an increased commercialization of VR devices, panoramic images and 360\degree\ videos have received significant attention in research into how to represent these spherical images for display in computer graphics\cite{chen2018recent, federico360video, hussain2021evaluation}. This has included advances into the various ways of representing spherical images in existing 2D formats. 
Such representations are not new, for they have been used by painters and in the cartography space for a while\cite{flocon1989curvilinear}, where map projections are used to represent the globe on a 2D surface\cite{snyder1997flattening}. 

Different projections will exhibit different distortions \cite{azevedo2019visual}. Projections can be broadly classified into viewpoint dependent projections, which depend on the user's view, and viewpoint independent projections, which are traditionally used in other areas like maps\cite{chen2018recent}. In this paper, we explore viewpoint independent projections for evaluating generated spherical images, focusing on the equirectangular and cubemap projections. Equirectangular panoramas are the most commonly used format in spherical image datasets, in part because they provide a 2D representation of the full content of the sphere as a single image with a 2:1 aspect ratio.
As an alternative we consider cubemaps, in which spherical image data is mapped to the faces of a cube surface. Cubemaps are already widely used in e.g. reflection and shadow mapping\cite{scherzer2011survey}, dynamic environment illumination\cite{ho2009unicube}, planet-sized terrain rendering\cite{dimitrijevic2016comparison}, and procedural textures\cite{zucker2018cube}. 
For graphics pipelines, other spherical representations will often be converted to a cubemap format before efficient rendering on a GPU. 

Projections onto polyhedra with a larger number of faces can reduce distortion, but potentially increases other issues like the number of seams.
We utilize this property of less distortion in polyhedron based representations in order to better employ FID on spherical images. 
Using projections for increasing compatibility of 2D image pretrained models with spherical images has previously been explored in other works like \cite{eder2020tangent} for semantic segmentation. 

\mypara{Image fidelity evaluation in generative models}
Fréchet Inception Distance (FID) is a widely established metric often used to measure image fidelity for evaluating image generative models, in part due to some agreement with human perception and sensitivity to various noise types \cite{Heusel2017GANsTB}. Under the assumption that this extends to equirectangular projections of spherical images, a majority of works in generative spherical imagery employ FID on this 2D representation as the main performance metric to measure the quality of generated images \cite{chen2022text2light, Lu2023AutoregressiveOO, wang2023360, akimoto2022diverse}. 
However, unlike regular images, 2D representations of spherical images must satisfy unique geometric constraints. 
We showcase the shortcomings of FID in evaluating geometric fidelity of spherical images, and we present an extension of the metric enabling a more efficient evaluation designed for spherical images by leveraging projections of spherical images. As such, our paper is an addition to prior works like \cite{naeem2020reliable, borji2022pros, chong2020effectively, Jayasumana2023RethinkingFT} that detect and tackle issues with FID.

\section{\oursFull\ Evaluation Metric}

Previous works in panoramic scene generation like \cite{chen2022text2light, Lu2023AutoregressiveOO, wang2023360, akimoto2022diverse} commonly use FID \cite{Heusel2017GANsTB} for quantitatively evaluating image generation quality. The FID score is computed between two sets of images, typically to assess the quality of a generative model by comparing the distance between the training set distribution and a corresponding generated distribution. All images from both image distributions are passed through the Inception V3 \cite{szegedy2016rethinking} convolutional model to obtain 2048-dimensional feature vectors. For each dataset, the obtained feature vectors are assumed to follow a multivariate Gaussian distribution with mean $\mu$ and covariance matrix $\Sigma$. The distance between the two distributions is then calculated using the Wasserstein-2 distance in $\mathbb{R}^{2048}$ \cite{Heusel2017GANsTB}, i.e. as 
\begin{align}
    FID(X_1,X_2) &:= d_{W-2}\left( \mathcal{N}\left(\mu_1, \Sigma_1\right), \mathcal{N}\left(\mu_2, \Sigma_2\right)\right) \\
    &= \Vert \mu_1 - \mu_2 \Vert + \text{tr}\left(\Sigma_1 + \Sigma_2 - 2 \left(\Sigma_1\Sigma_2\right)^{\frac{1}{2}} \right)
\end{align}
Although the underlying Gaussian assumptions have been shown to be faulty \cite{luzi2023evaluating}, FID has been established as a popular metric due to sensitivity to noise and some correlation with human perception \cite{Heusel2017GANsTB}. 

Notably, however, spherical images present additional geometric structure compared to regular 2D images. 
It is not clear a priori whether the features produced by the Inception backbone, and hence the FID metric by extension, will capture divergences from these geometric constraints, even when the ground truth reference set contains proper projections of spherical images. Indeed, local image properties may well look reasonable, but global information in 2D representations is required to assess whether the geometric constraints are fulfilled (e.g. seamless stitching at poles and across image borders). 
One option to address this is adaptation of the Inception network to spherical images. Possibilities include adapted sampling strategies for the convolution using projections as in \cite{Coors2018SphereNetLS, Tateno2018DistortionAwareCF, Eder2019ConvolutionsOS}, either combining with the network in a zero-shot sense, or training a new such model on spherical image data to replace the Inception network. However, the sustained popularity of FID can ultimately be attributed to useful and robust Inception features, and these options will alter the features in an unclear way. Thus, with considerations of the proven record of FID, rather than tampering with the metric or underlying model, our strategy is to instead adapt the spherical images to the Inception network.

\begin{figure}[t]
  \centering
  \includegraphics[width=\textwidth]{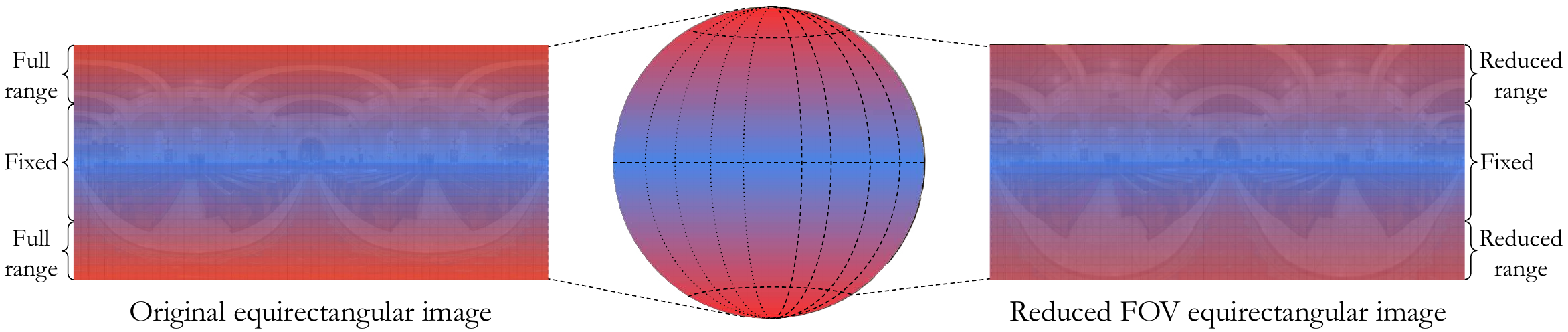}
  \caption{Visualisation of the noise transformation used for reducing field-of-view in spherical images. The proposed  transformation reduces vertical field-of-view while maintaining proportions in the central horizontal parts of the equirectangular image. 
  }
  \label{fig:fov-noise}
\end{figure}

\mypara{Increasing compatibility of FID to spherical images} In order to improve conformity of the spherical images to the Inception backbone of the FID metric, we utilize the commonly used tangential spherical cubemap projection as grounds for evaluation. Further, since the resulting views are square, the aspect ratio is maintained when resizing the inputs for the network to the expected $299 \times 299$ pixels, in contrast to using equirectangular images. 

As transformations between projections will incur some image quality degradation \cite{hanhart2018360}, we believe it crucial to evaluate image fidelity on representations that are optimized for rendering for a representative evaluation. Further, since hardware and shaders have been optimized both for equirectangular and cubemap projections, evaluation on cubemap projections is not only valid, but perhaps even desirable. 
We note that although we focus on evaluation on the tangential spherical cubemap in this work, fair comparisons are also possible on other cubemap representations and re-projections, as long as the representations are unified. This could be relevant if generative models are to be evaluated for a specific shader or other transforms of the representations.

For a spherical image dataset $X$, we denote the set of 2D images resulting from cubemap projections by
\begin{equation}
    \mathcal{C}^X := \lbrace \mathcal{C}^X_F, \mathcal{C}^X_R, \mathcal{C}^X_B, \mathcal{C}^X_L, \mathcal{C}^X_U, \mathcal{C}^X_D \rbrace , 
\end{equation}
with the resulting view-specific image sets being denoted as $\mathcal{C}^X_{view}$, where $F$, $R$, $B$, $L$, $U$, and $D$ represent the front, right, back, left, up, and down view of the cubemap projections, respectively. With the notation above, we focus on the set structure of the individual cubemap views. 

A priori we expect that the Inception feature distributions across cubemap views will differ. Concretely, we hypothesize that the feature vectors of the frontal views (front/right/left/back) are identically distributed, since the orientation of these views are arbitrary, but that the up and down view feature distributions will be dissimilar. Properties of the tangential spherical cubemap projection additionally support this, since structural distortions are larger at the polar faces (upwards and downward) compared to frontal faces, as a results of stitching at the poles. To get empirical evidence for this we compare the feature means of the different views on the 360-Indoor dataset \cite{Chou2019360IndoorTL}. Between any two frontal views, the $L^2$ distances between mean features are $0.34 \pm 0.13$, while it is  $24.27 \pm 0.69$ and  $33.13 \pm 0.59$ between frontal views and up/down views, respectively. Additionally, the $L^2$ distance is $17.05$ between the average features of up and down views. 

\mypara{\ours} On this basis, we group the cubemap views $\mathcal{C}^X$ into three disjoint subsets according to semantic similarities between frontal views $\mathcal{F}:=\lbrace F, R, B, L \rbrace$, upward views $U$ and downward views $D$. The frontal group $\mathcal{F}$ consists of four times as many perspective images as $U$ and $D$. 
Since FID is biased by sample size \cite{chong2020effectively}, for every cubemap we first average the Inception features of the perspective images within each view group ($\mathcal{F}$, $U$, and $D$) before computing FID on the resulting sets of features. We denote this by $\overline{FID}$. 
Averaging these scores, we get our proposed extension \ours:
\begin{equation}
    \ours(X_1, X_2) := \frac{1}{3} \sum_{V\in\lbrace U, D, \mathcal{F}\rbrace} \overline{FID}(\mathcal{C}^{X_1}_V, \mathcal{C}^{X_2}_V)
\end{equation}

\begin{figure}[t]
  \centering
  \begin{subfigure}{0.49\linewidth}
    \centering
    \includegraphics[width=0.9\linewidth]{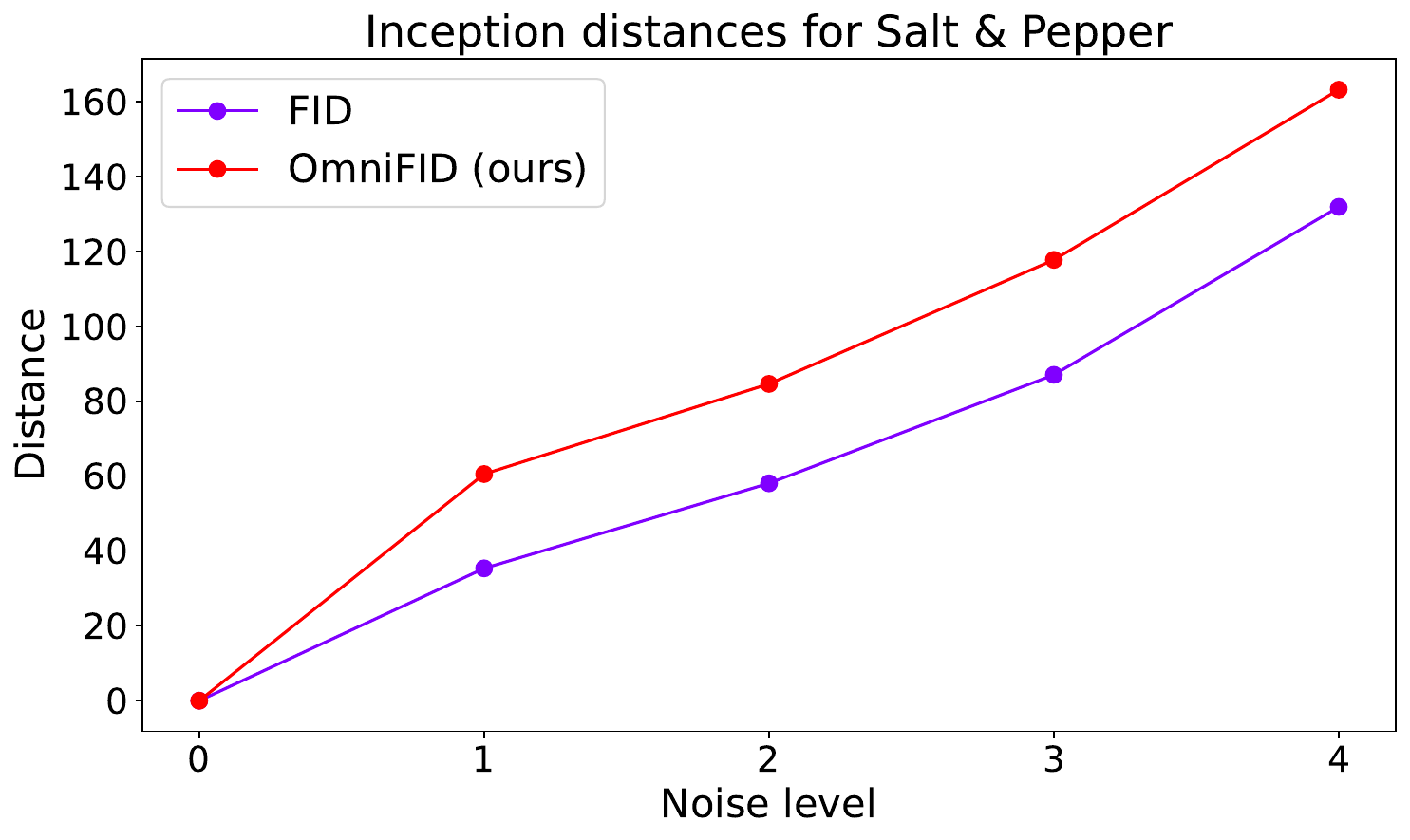}
    \label{fig:upscale-b}
  \end{subfigure}
  \begin{subfigure}{0.5\linewidth}
    \centering
    \includegraphics[width=0.49\linewidth]{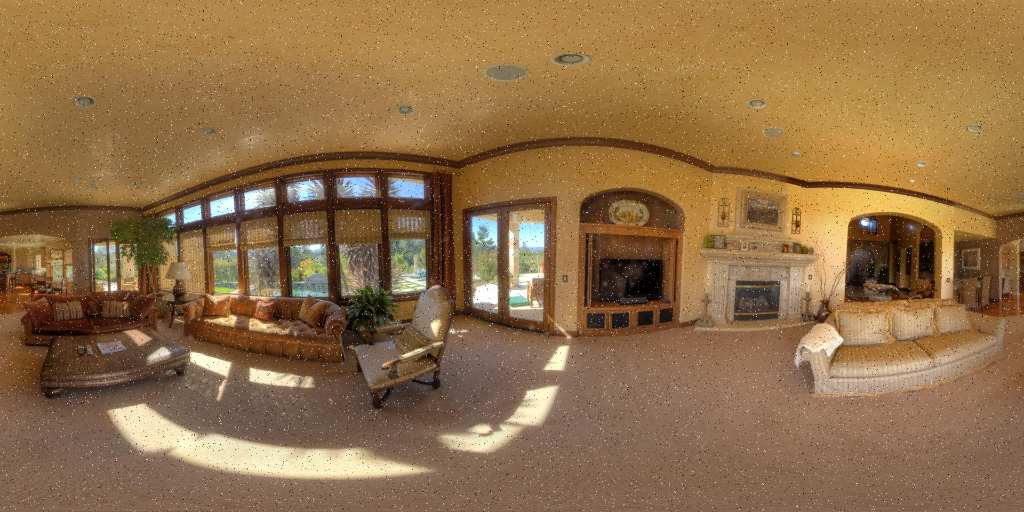}
    \includegraphics[width=0.49\linewidth]{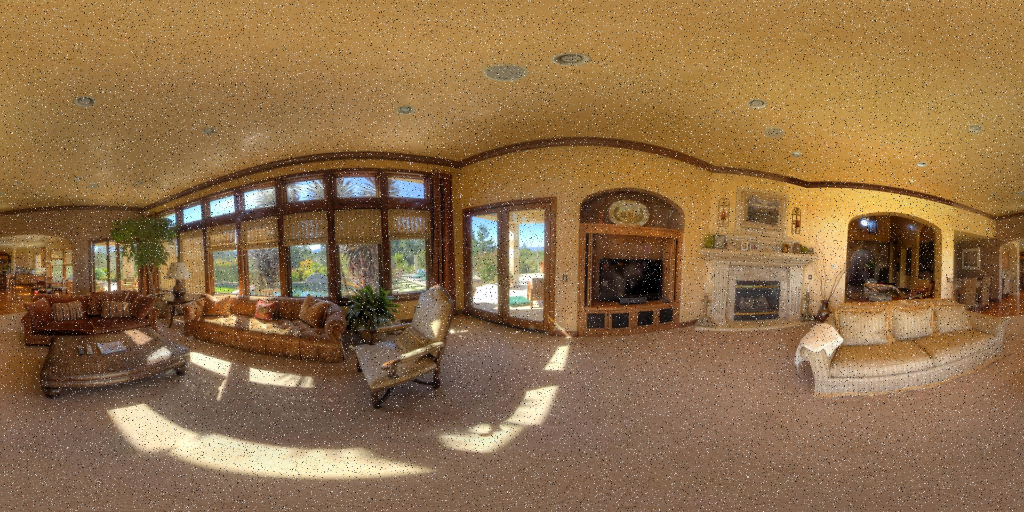}
    \includegraphics[width=0.49\linewidth]{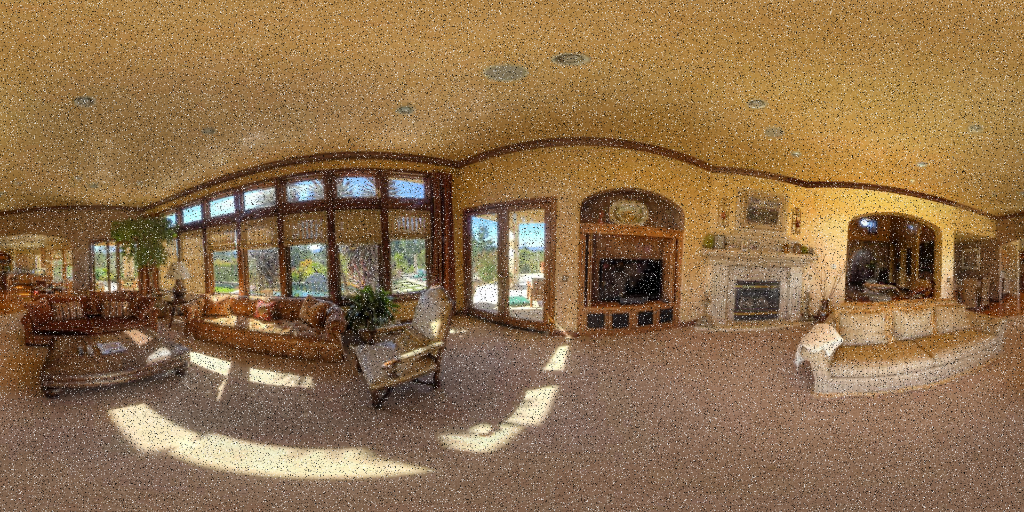}
    \includegraphics[width=0.49\linewidth]{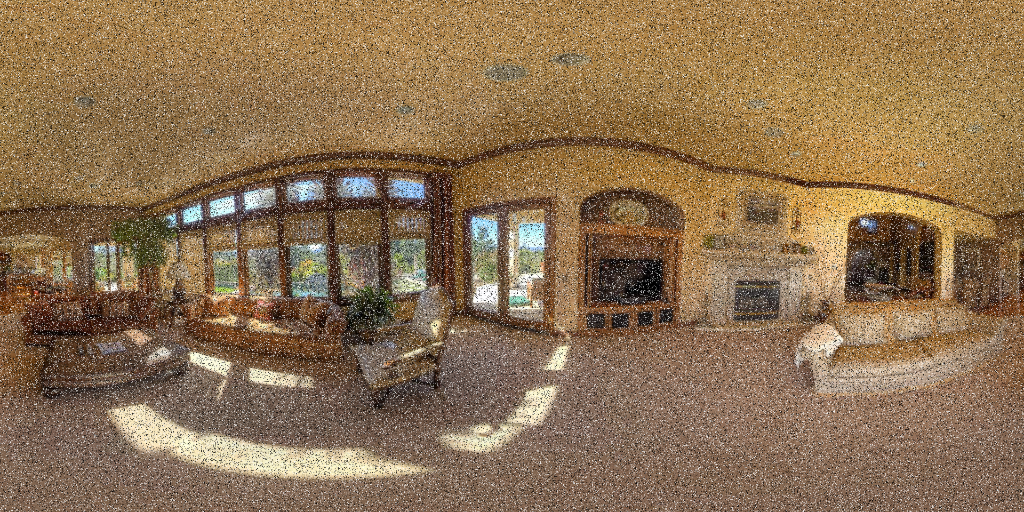}
    \label{fig:upscale-b}
  \end{subfigure}
  \begin{subfigure}{0.49\linewidth}
    \centering
    \includegraphics[width=0.9\linewidth]{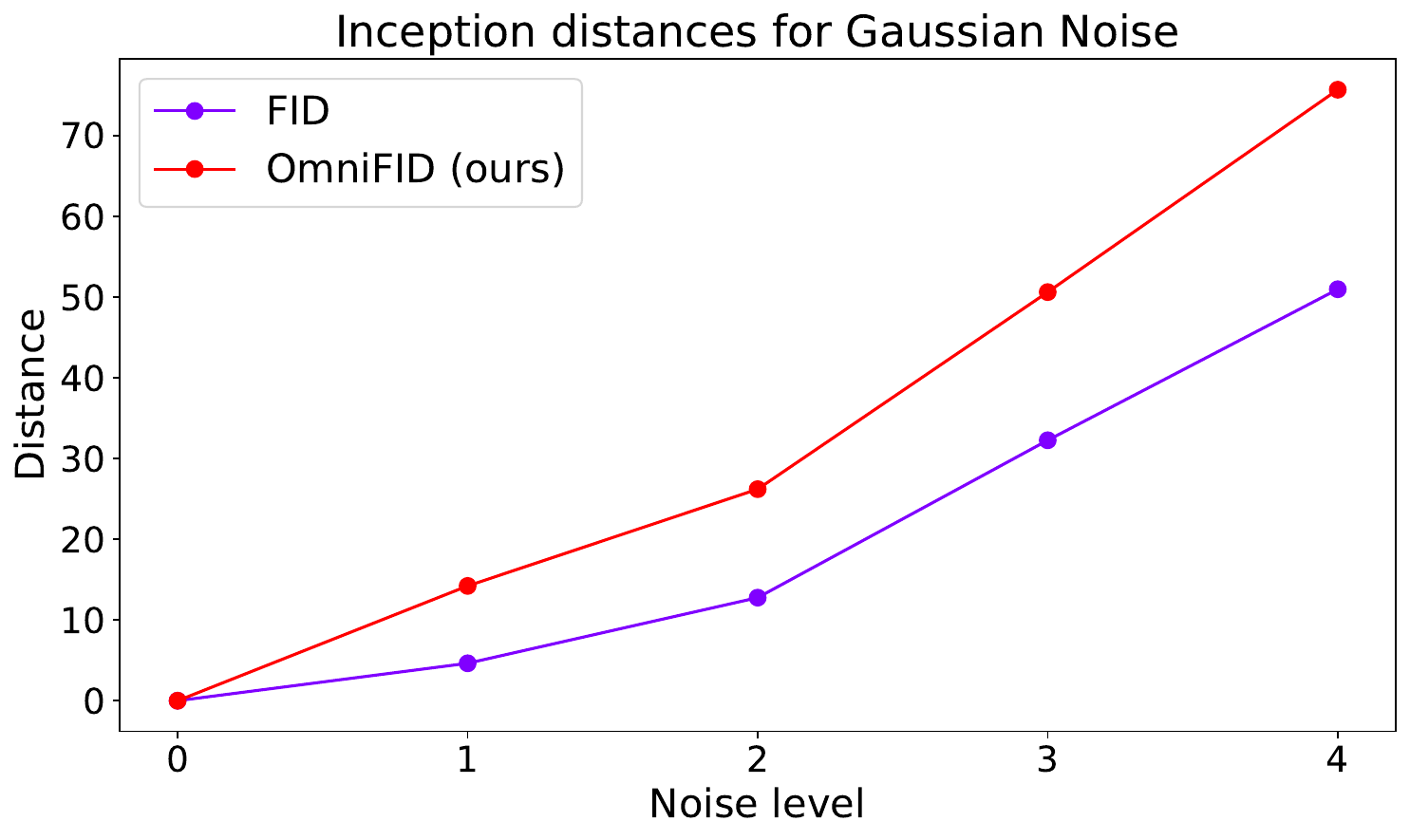}
    \label{fig:upscale-b}
  \end{subfigure}
  \begin{subfigure}{0.5\linewidth}
    \centering
    \includegraphics[width=0.49\linewidth]{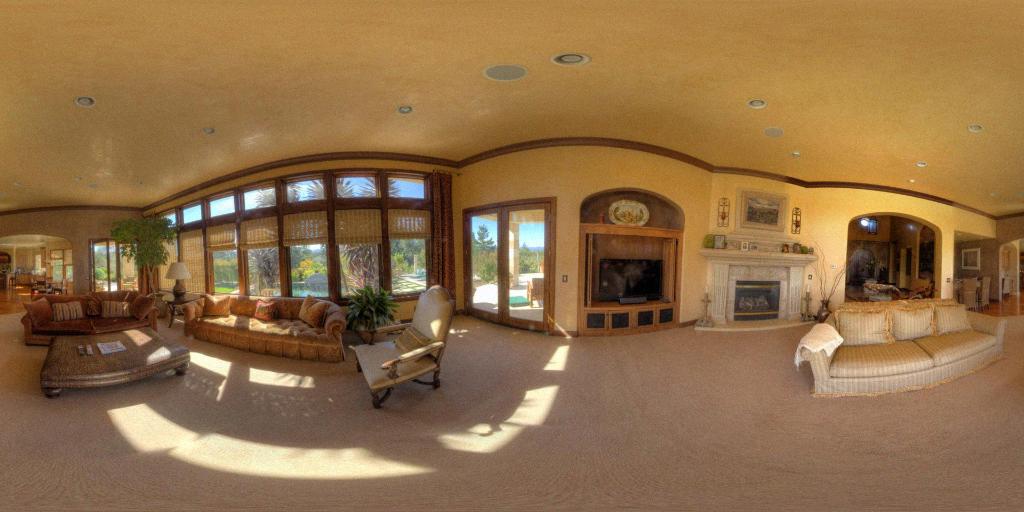}
    \includegraphics[width=0.49\linewidth]{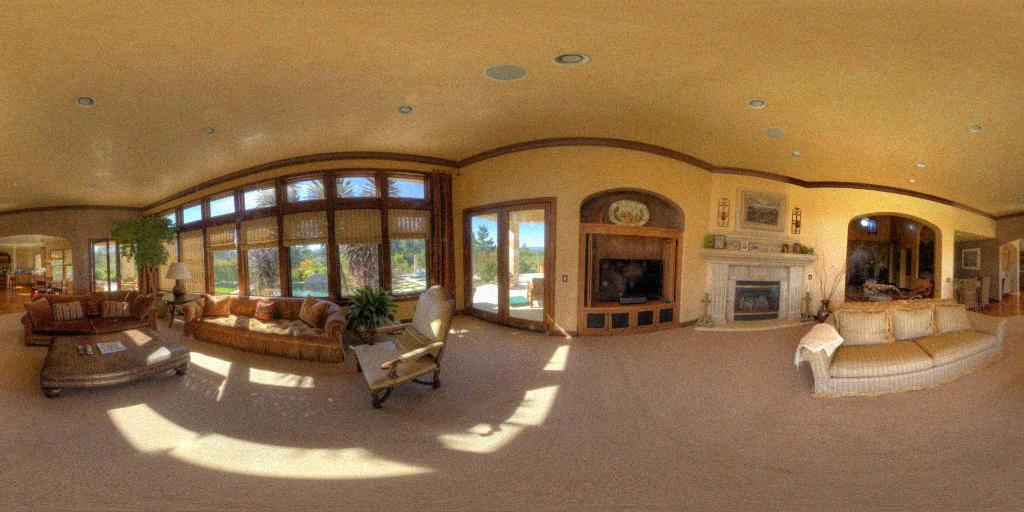}
    \includegraphics[width=0.49\linewidth]{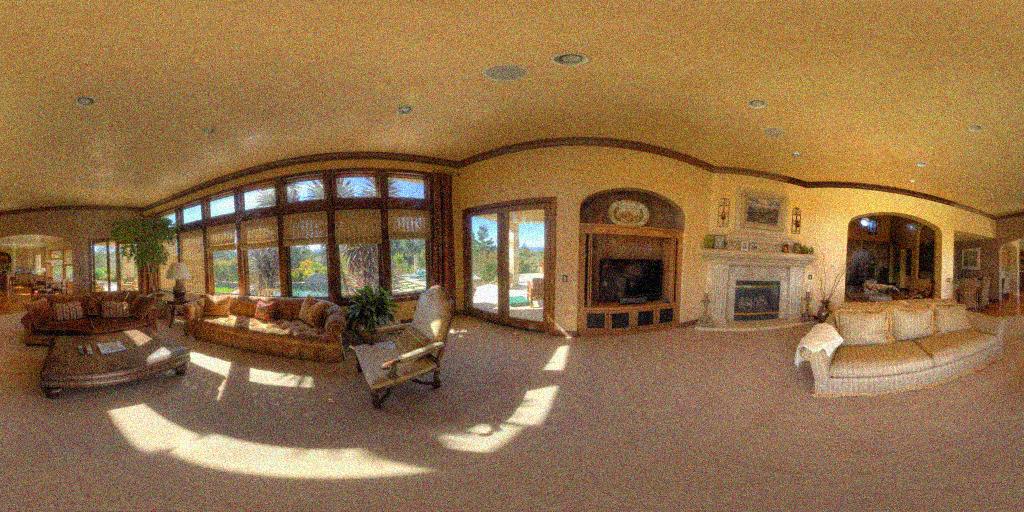}
    \includegraphics[width=0.49\linewidth]{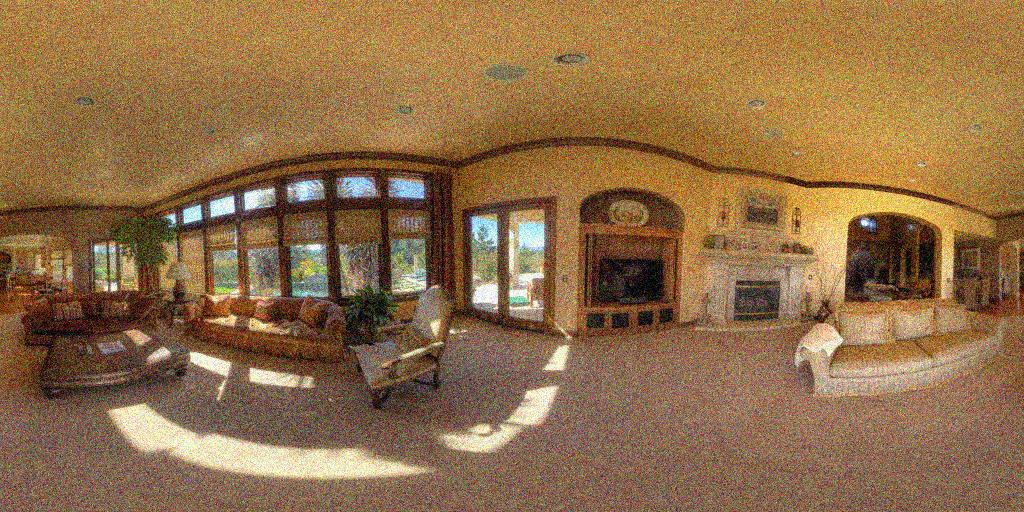}
    \label{fig:upscale-b}
  \end{subfigure}
  \begin{subfigure}{0.49\linewidth}
    \centering
    \includegraphics[width=0.9\linewidth]{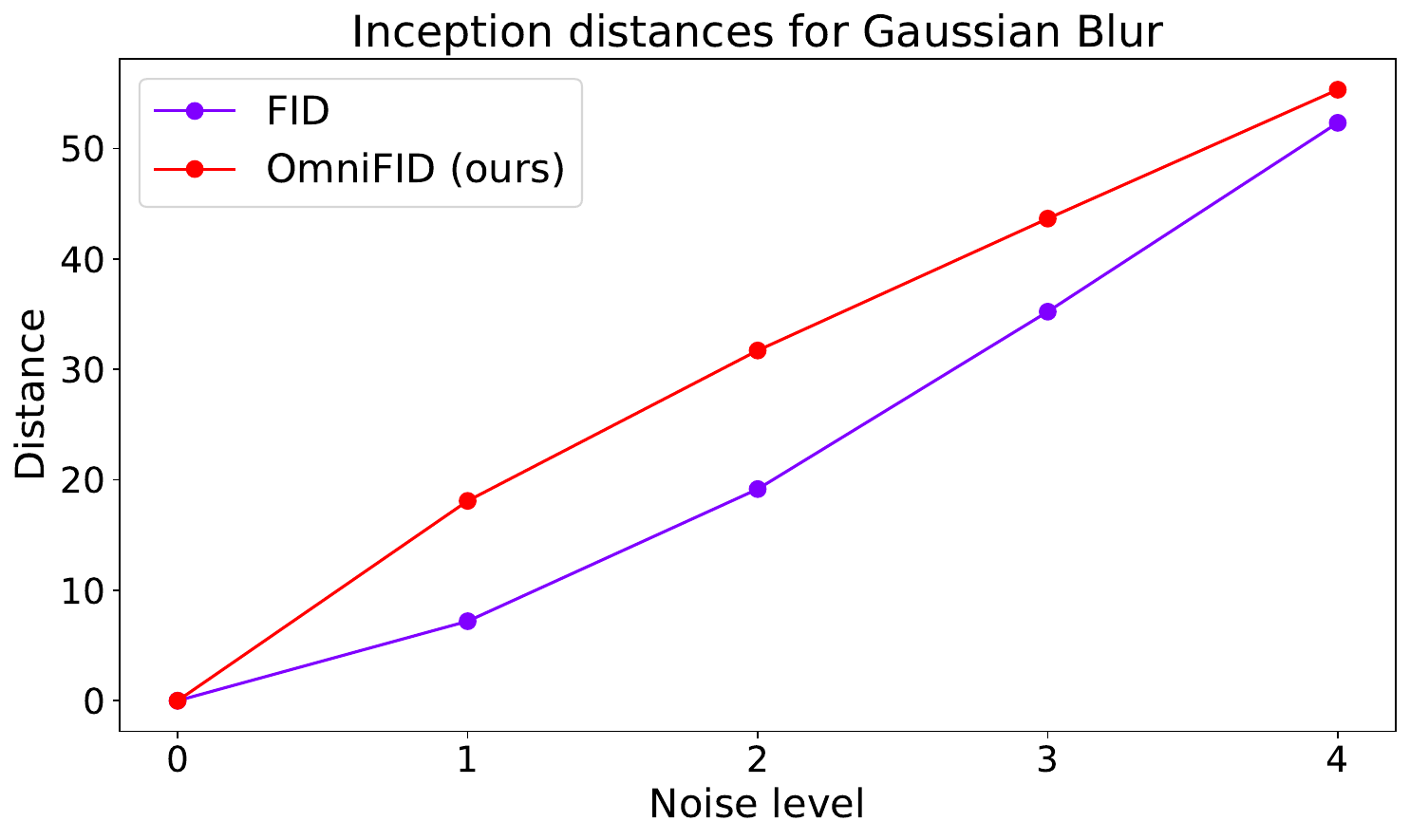}
    \label{fig:upscale-b}
  \end{subfigure}
  \begin{subfigure}{0.5\linewidth}
    \centering
    \includegraphics[width=0.49\linewidth]{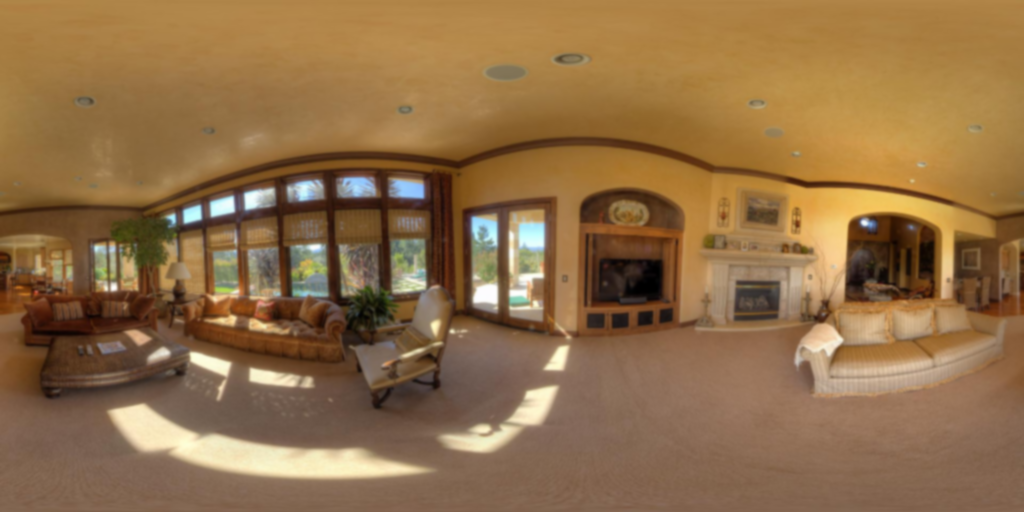}
    \includegraphics[width=0.49\linewidth]{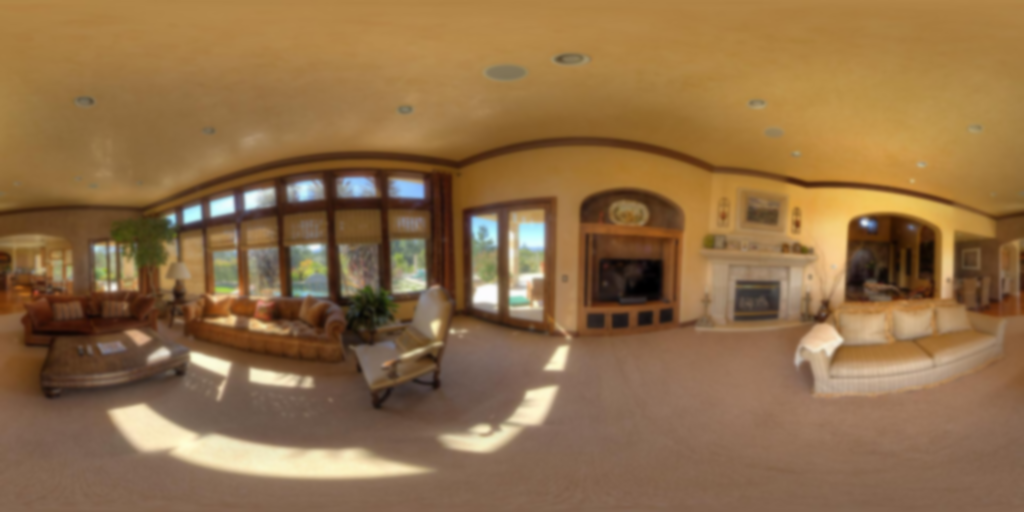}
    \includegraphics[width=0.49\linewidth]{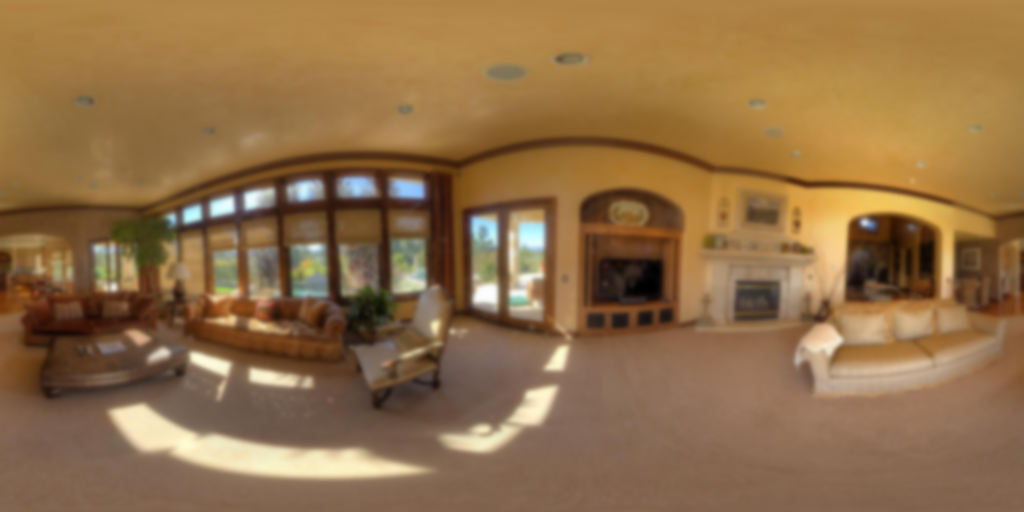}
    \includegraphics[width=0.49\linewidth]{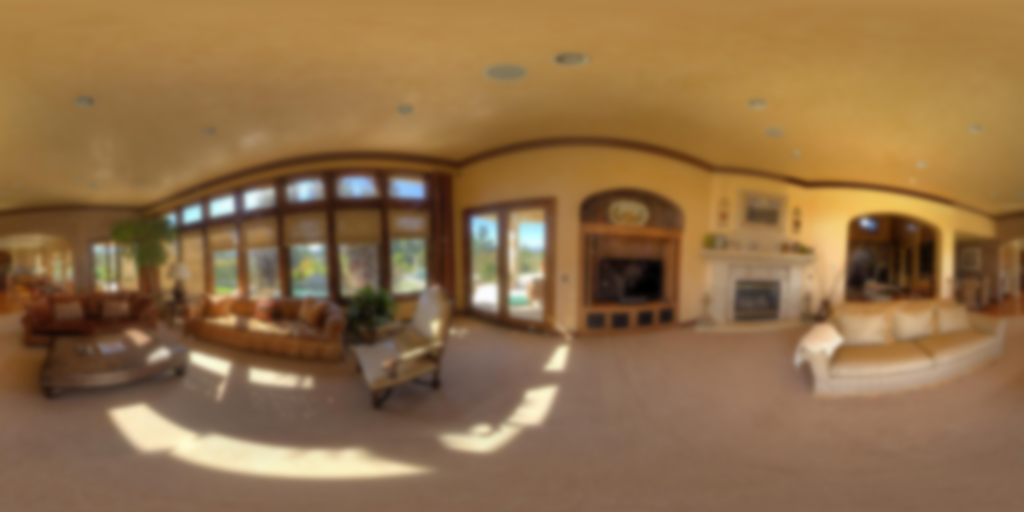}
    \label{fig:upscale-b}
  \end{subfigure}

  \caption{FID compared to our \ours\ on various noise types unrelated to the spherical geometry (salt \& pepper, Gaussian noise, and Gaussian blur). Noise is applied to the equirectangular image, before it is transformed to the cubemap. \ours\ retains the noise-related properties of FID.  
  }
  \label{fig:omnifid-noise-types}
\end{figure}

In \Cref{sec:exp} we show that using cubemaps is adequate for detecting structural issues related to field-of-view in spherical images, while benefiting from ease of implementation and limited additional compute. We also test that \ours\ maintains desirable properties of the ordinary FID metric. In principle however, \ours\ is independent from the underlying partitioning of the sphere, and we discuss how \ours\ can be extended to other projections in \Cref{sec:lim-further}. 

\section{Discontinuity score}
\begin{figure}[t]
  \centering
  \includegraphics[width=0.49\textwidth]{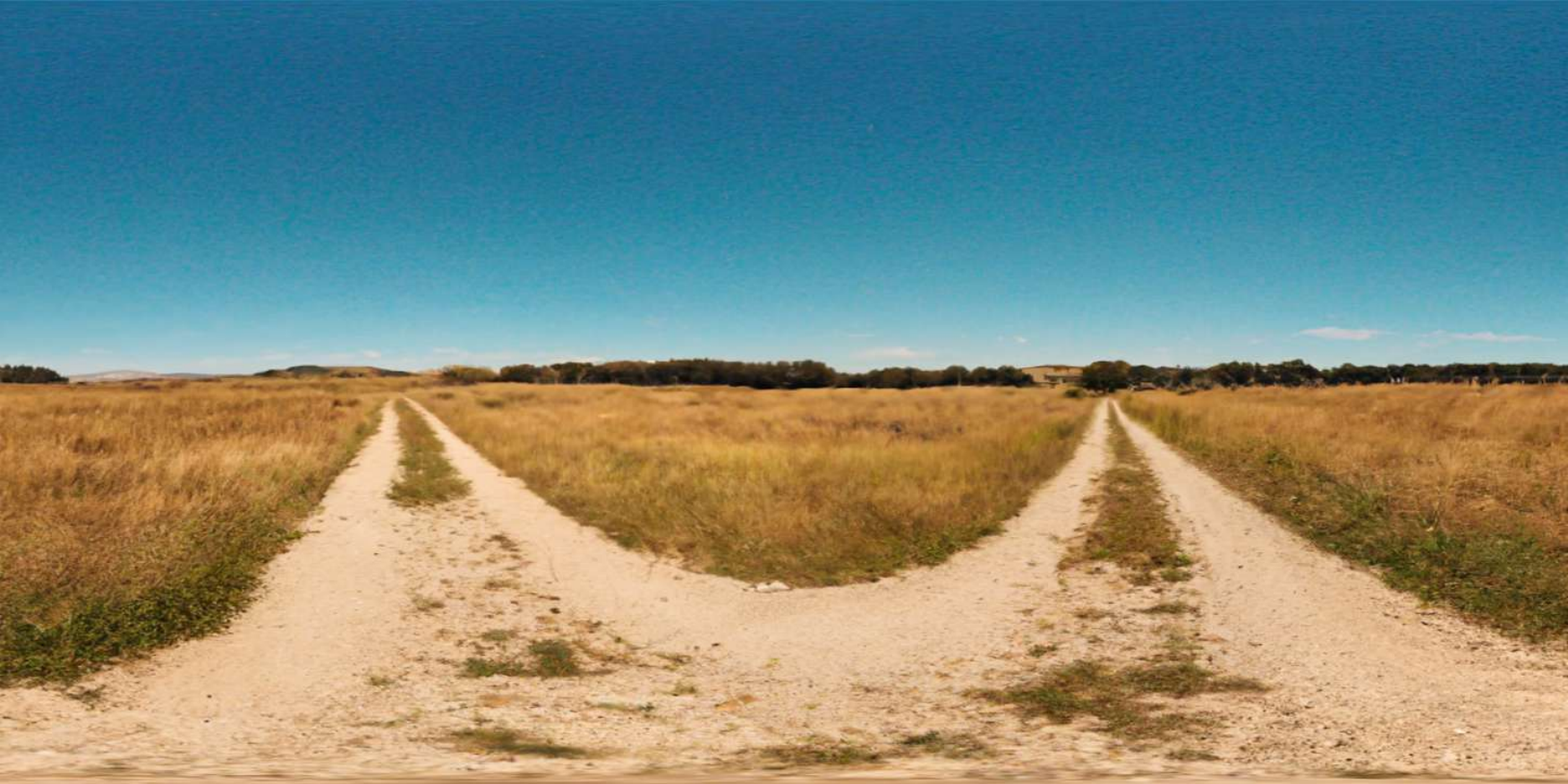}
  \includegraphics[width=0.49\textwidth]{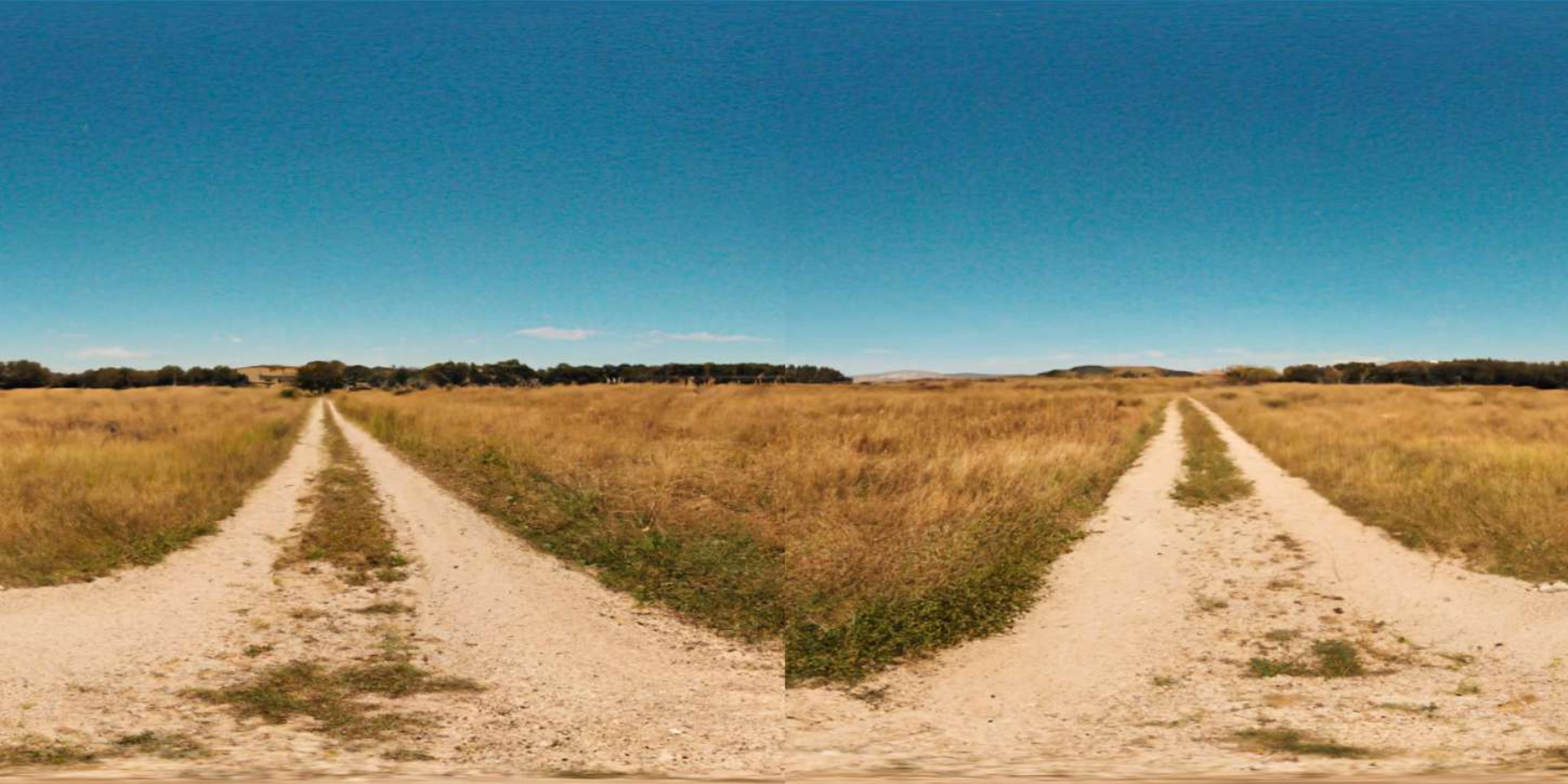}
  \caption{Example of discontinuities resulting from seam issues in generated images, i.e. that image borders do not properly align. The right image is equal to the left image, but rotated 180\degree\ horizontally. This is a common issue when spherical images are generated via intermediate 2D image representations like equirectangular or cubemap images. Both FID and \ours\ fail to detect such issues, which leads us to define our Discontinuity Score (DS).
  }
  \label{fig:seam-example}
\end{figure}

Another requirement of planar representations of spherical images is that image borders must align when projected back to the sphere in terms of both semantics, lighting etc. If such misalignments exists, rendering of the spherical image will contain a visible line separating two areas of the sphere that are incompatible (see e.g. \Cref{fig:seam-example}). Adhering to these geometric constraints is important for the rendering to feel natural, and as such it should be an important aspect of spherical image and model evaluation that such constraints are taken into account. In this section, we present Discontinuity Score (DS), a simple kernel-based edge detection algorithm for evaluating image border alignment issues in individual equirectangular images resulting in visible seams, or interrupts.

\mypara{The DS algorithm} For a given generative model, it is known a priori which 2D image representation the output image will have (e.g. equirectangular, cubemap or something else). As such, the locations of possible seams are known, which we utilize to score seam issues without needing to detect and distinguish between semantically correct edges and edges resulting from image border discontinuities.   
We can therefore isolate potential seams, indexed by $i$, that may exist across borders in the image $I$. For each potential seam in $I$, we create an array of pixels $a_i$ surrounding the potential seam with height equal to the potential seam length $L$, and a width of 6 pixels. The array $a_i$ is then converted to greyscale. Since we want to quantify how abruptly pixel intensities change exactly when crossing the potential seam, a small $3 \times 3$ kernel is used for horizontal edge detection. We follow the recommendation of using a Scharr kernel $K$, since it gives a better approximation of the derivative when using  a $3 \times 3$ kernel, as is common practice in libraries like OpenCV \cite{opencv_library}.

The simplest way to construct a single scalar score for each such array $a_i(x,y)$, would be to compute the convolution $\hat{a}_i(x,y) = K * a_i(x,y)$, and average the values along the seams, i.e. $\frac{1}{2L}\sum_{x\in\lbrace 2,3 \rbrace}\sum_{y=0}^{L-1} \left|\hat{a}_i(x,y)\right|$ using zero-indexing. However, we find that this formulation of the score does not adequately score seams with abrupt semantic stops such as objects disappearing across the seam, and is affected by surfaces with clutter. 
To increase the impact of abrupt semantic discontinuities, we instead consider the relative change from one side of the seam to the other. Concretely, we compute the array-wise scores by
\begin{equation}
    DS(a) :=  \frac{1}{2L}\sum_{y=0}^{L-1} \left( \frac{\left|\hat{a}(2,y)\right|}{\left|\hat{a}(1,y)\right| + c} +  \frac{\left|\hat{a}(3,y)\right|}{\left|\hat{a}(4,y)\right| + c} \right) ,
\end{equation}
where $c$ is a constant introduced to stabilize the score and avoid division with zero. Finally, we get the complete image-wise DS value by summing over the seam scores, accounting for the seam length $L$ relative to the height of the corresponding equirectangular image $H_E$:
\begin{equation}
    DS(I) := \frac{L}{H_E}\sum_i DS(a_i) .
\end{equation}
We do not average over the number of arrays in an effort to make the score adjust to different 2D representations of spherical images, i.e. a representation with more seams should be able to get a comparatively higher DS score. 

\section{Experiments}
\label{sec:exp}
\subsection{OmniFID Evaluation}
\label{ssec:fid-eval}
\mypara{Field-of-view reduction noise} In order to evaluate the ability of FID to capture issues related to the geometric requirements of spherical images, we construct a noise transformation for reducing the vertical field-of-view in equirectangular projections of spherical images. The noise transformation is visualized in \Cref{fig:fov-noise}, and the effects, which are a common artifact in generated equirectangular images, can be seen in \Cref{fig:fov-issues-example}, along with realistic examples from the literature in \Cref{fig:sota-issues}. Concretely, the field-of-view is reduced by an angle $v$ by first cropping the top and bottom horizontal parts of the equirectangular image corresponding to $\frac{1}{2}v$ each. The central 90\degree\ horizontal part of the image is kept fixed, while the remaining parts of the image, each covering $90\degree-\frac{1}{2}v$, are resized using bi-linear interpolation to re-obtain the original image resolution. 

For our evaluations of FID and \ours, we use the 360-Indoor dataset \cite{Chou2019360IndoorTL}, a collection of 3335 equirectangular images of indoor scenes with 360\degree\ horizontal and 180\degree\ vertical field-of-view. The images have resolution $1920 \times 960$, and we resize them to $1024 \times 512$. Compared to other datasets of spherical images, 360-Indoor is optimal for this purpose since it has both full field-of-view and has enough samples for the mean and covariance estimates in the FID and \ours\ metrics to be valid, although the number of samples is still relatively low. In the experiments, we use an uncorrupted copy of the 360-Indoor dataset, and a copy which we gradually corrupt - here with reduction of vertical field-of-view. 

In \Cref{fig:exp-fov-noise}, we see that decreasing the field-of-view from 180\degree\ to 140\degree\ results in an FID of just 10. This is a particularly low value considering that the bias of FID depends on sample size, since the 360-Indoor dataset contains just 3335 spherical images \cite{chong2020effectively}. Further, when comparing with FID values resulting from other types of noise (as shown in \Cref{fig:omnifid-noise-types}), it is also evident that FID captures this geometric issue insufficiently.
On the other hand, \ours\ crucially captures the difference in geometric fidelity between the real and corrupted dataset. This confirms that while FID fails to capture important aspects of the quality of spherical images, the adjustments made in \ours\ allows the metric to better quantify fidelity related to vertical field-of-view.

\begin{figure}[t]
  \centering
  \includegraphics[width=0.99\textwidth]{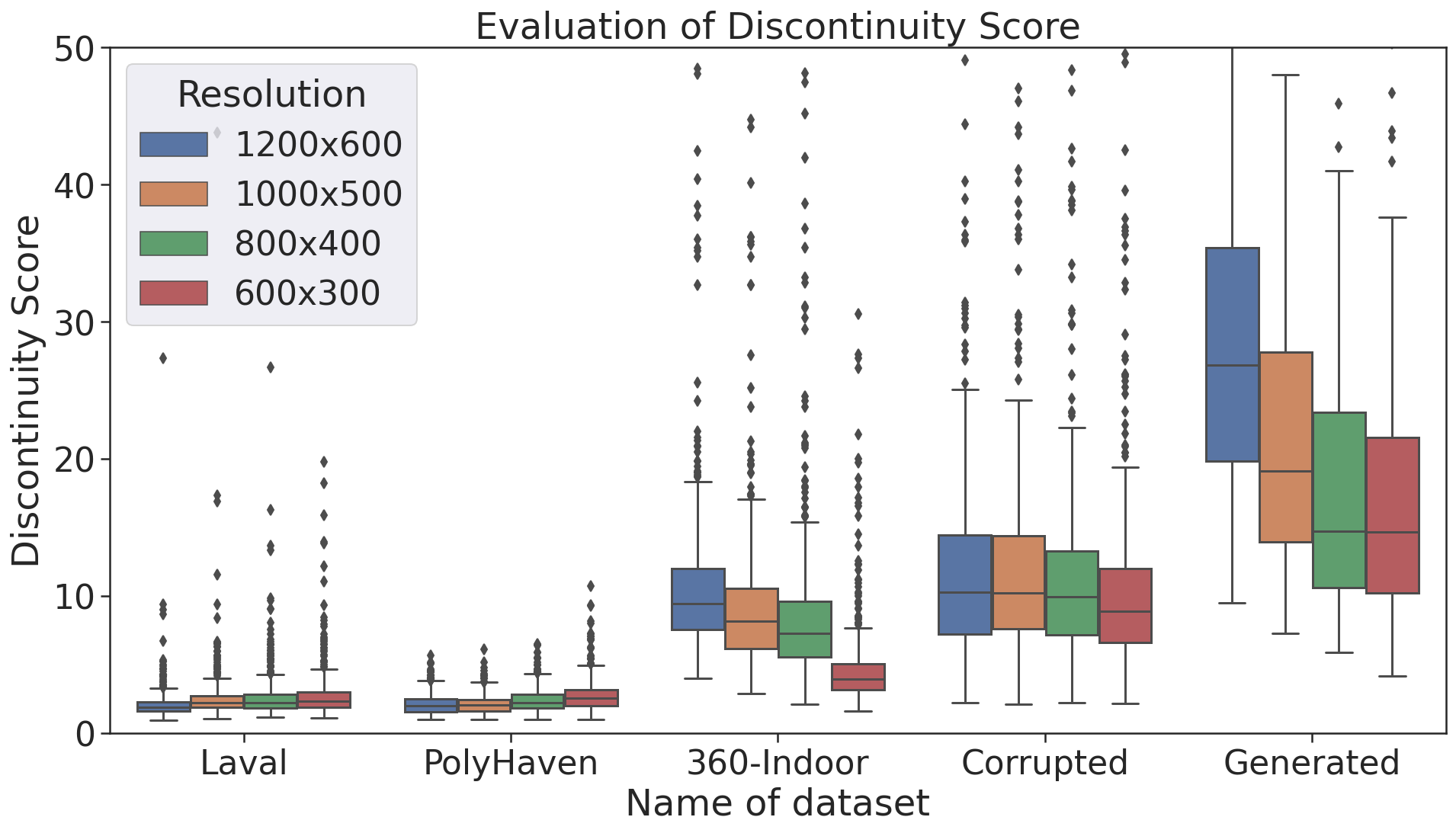}
  \caption{Comparison of Discontinuity Score (DS) on datasets with and without seam issues across various image sizes. 360-Indoor \cite{Chou2019360IndoorTL}, Polyhaven (indoor and outdoor) \cite{polycitat}, and Laval \cite{gardner2017learning} are spherical image datasets. 360-Indoor contains very mild seam issues. "Corrupted" denotes images from the Laval dataset with pixels cropped from left and right border to create artificial mild interrupts. "Generated" contains spherical images with semantic interrupts as an artifact from the generative process. Conclusions are consistent across image resolutions. We limit the y-axis to 50 for better visualization.
  }
  \label{fig:ds-robustness}
\end{figure}

\mypara{Additional noise evaluations} The FID metric became an established metric in part due to its sensitivity to various forms of noise \cite{Heusel2017GANsTB}. Here, we validate that \ours\ has not lost these properties of the FID metric through our extension. As above, we use two copies of the 360-Indoor dataset, gradually corrupting one with salt \& pepper noise, Gaussian noise, and Gaussian blurring, respectively. We then compute FID and \ours\ between the two dataset. For each type of corruption, we increase the noise over four levels of noise strengths. We note that the noise is applied on the equirectangular image, i.e. before transforming the images to cubemaps for \ours. Results are visualized in \Cref{fig:omnifid-noise-types}, along with example equirectangular images showing the level of noise. We observe that for the different noise types, \ours\ follows the trend of FID closely, demonstrating that our extension retains these desired properties of FID.  We further note that similarities between the FID and OmniFID scores across noise types and levels confirm that the difference in scores on the field-of-view reduction task are not simply a matter of scaling. 

\mypara{Qualitative evaluation of \ours} In the \supp we compare \ours\ and FID scores on generated examples of varying quality from different checkpoints of a generative model based on Imagen \cite{saharia2022photorealistic}, trained on internal data sources and finetuned using dreambooth on the 360-Indoor dataset \cite{ruiz2023dreambooth}. We showcase that \ours\ decreases as adherence of generated images to the spherical structure improves, while FID is unaffected - in fact, the lowest of the FID scores is achieved on a set of images with clear geometric issues. We note that similarity of features between frontal views is also present on the generated dataset, serving as further motivation for the Discontinuity Score.

\begin{figure}[t]
  \centering
  \includegraphics[height=6.1cm]{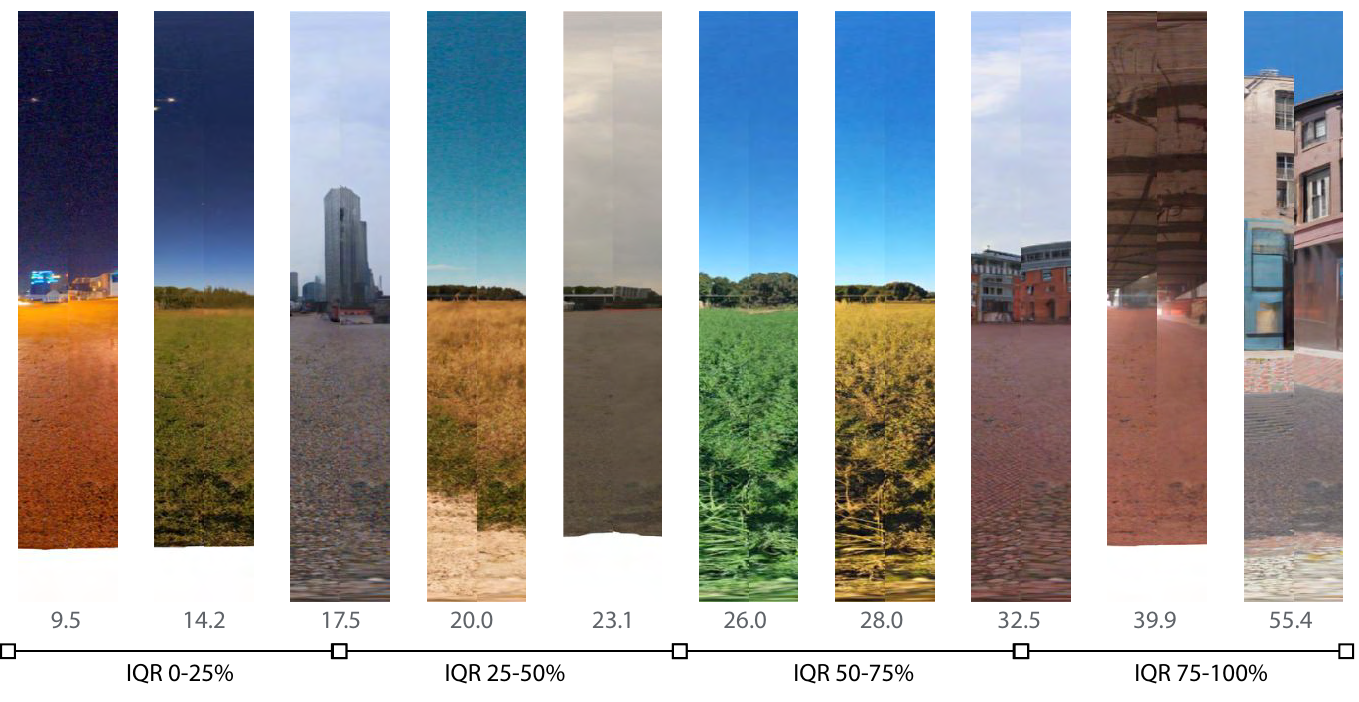}
  \caption{Qualitative evaluation of the discontinuity score on generated spherical images with interrupts across image borders. From a generated dataset, each image is scored using the proposed DS score. Here, we visualize the seams of the images corresponding to increments of 10th percentiles of DS. Larger DS aligns well with perceived inconsistencies (e.g. in semantics and lighting) across the seam.  We mark interquartile ranges (IQR) for visualization purposes.
  }
  \label{fig:ds-qualitative}
\end{figure}

\subsection{Evaluation of DS}
\label{ssec:ds-eval}
\mypara{Robustness of DS} We compare computed scores on three spherical image datasets, a corrupted dataset, and a generated dataset with seam issues in \Cref{fig:ds-robustness}. Scores are computed for different resolutions to validate that conclusions made by DS are robust across image sizes. We use three datasets with equirectangular images, namely 360-Indoor \cite{Chou2019360IndoorTL}, PolyHaven \cite{polycitat} and Laval \cite{gardner2017learning}. For our corrupted dataset, we crop pixels from left and right image borders in the Laval dataset corresponding to $0.25\%$ of the image width in each side. This creates a small discontinuity of mild semantic nature. For our generated dataset, we use a generative model based on Imagen \cite{saharia2022photorealistic}, trained on internal data sources and finetuned using dreambooth on equirectangular images from PolyHaven and Laval \cite{ruiz2023dreambooth}. We sample a dataset of 143 equirectangular images from the model. 
\Cref{fig:ds-qualitative} shows a subset of generated images with various degree of discontinuities across image borders, demonstrating a good use-case of DS. 
Empirically, we found that the second-order Scharr kernel and scalar hyperparameter $c=0.1$ provided good results, and we use these settings across all experiments. 

In \Cref{fig:ds-robustness} we observe that the values of DS reflects the level of seam misalignment embedded across different datastets enabling detection of images with seam issues apart from a few outliers. On the ground truth spherical image datasets, the mean score is near 0, except on 360-Indoor which does contain a very mild seam. Corrupting the seam of the Laval dataset leads to a substantial increase in score as expected. Finally, the generated dataset have the largest discontinuity scores, explained by some images having gross semantic misalignments.  
The results also suggest that the conclusions are consistent across the four different image resolutions. We attribute some of the differences in DS score across image resolution to anti-aliasing smoothing effects of the image resizing functions: indeed, on the corrupted datasets, where we create the artificial seam after resizing, DS scores are near-constant. 

\mypara{Qualitative of DS} 
In \Cref{fig:ds-qualitative}, we showcase the seams in each 10th percentile of the images in the generated dataset, ordered by their discontinuity score from left (lower) to right (higher). All images contain a visible seam, however, the severity of discontinuities in lighting and semantic content increases. As an example, in the second image, the grass, trees, and different layers of illumination of the sky are well-aligned, while in the right-most example, buildings stop abruptly across the image borders, and the direction of roads do not match. 

\section{Limitations and further research}
\label{sec:lim-further}
While our proposed metrics better reflect the unique properties of spherical images than current alternatives, we will here discuss some limitations and directions for future research.
Although cubemap representations facilitate the translation of prior knowledge in 2D image-based models and metrics to the spherical image domain, treating the faces of the cube individually leads to a loss of global information. This could lead to ignoring semantic inconsistency across seams to some degree. Our DS metric tries to address this shortcoming in part. 
On the other hand, although we have shown that using the six faces of the cubemaps makes it possible to detect field-of-view issues, it is unclear if it is necessary to further reduce projection distortion in order to sufficiently evaluate capabilities going forward, as access to high-quality spherical image datasets and generative models increases. We discuss possible adaptations of \ours\ to this scenario below. 
The current limited availability of spherical image datasets and spherical image generative models limits the scope of our experimental results to these ends, in particular for benchmarking existing generative models. Indeed, publicly available datasets like PolyHaven \cite{polycitat} and Laval \cite{gardner2017learning} come with their own challenges, such as small dataset sizes for valid mean and covariance estimates for computing (Omni)FID, as well as field-of-view less than 180\degree.  

\mypara{Cubemap alternatives} \ours\ can be computed with perspective images from finer partitionings of the sphere. In particular, icosahedron tangent images \cite{eder2020tangent} provide a fitting way to reduce distortion in the 2D perspective images at the cost of less semantic content in individual views, additional computation, and overlapping content between images. The tangent images can be grouped based on the latitude of their centers, as done with regular cubemaps above. As an example, a base level 0 icosahedron subdivision of the sphere gives 20 perspective images, which come in four groups of five images with centers at the same latitude. \ours$^{20}$, with the superscript denoting the number of perspective images, can then be computed on these images by first computing the Inception features on each image, averaging the features over each of the four groups, and computing the corresponding latitude-wise $\overline{FID}$ scores. Finally, the $\overline{FID}$ scores can be averaged as before. Other alternatives include (combinations of) projections with different properties, such as equiangular cubemaps \cite{equiangular}. 

\mypara{Perceptual metrics} An interesting question we leave for further research is how perceptual metrics like LPIPS \cite{zhang2018unreasonable} relying on convolutional networks are affected by different 2D projections and sampling of the sphere. Low-level perceptual metrics like SSIM \cite{Wang2004ImageQA} have previously been improved for spherical images \cite{Chen2018SphericalSS}.

\section{Conclusion}                         
In this work we showcase that the standard image fidelity metric FID, commonly used in evaluation of generative models, fails to capture crucial properties of spherical images associated with their unique geometrical constraints. To remedy the limitations of existing 2D image-based metrics, we presented an extension of FID, called OmniFID, and a discontinuity score to quantify the geometric distortion and seam misalignment across image borders, respectively. Experiments demonstrate the effectiveness of our proposed metrics to measure geometry fidelity for spherical images. Our contributions advance spherical image evaluation as immersive content generation for spatial computing devices is gaining traction. The work encourages further research by others, and provides avenues for metric adaptations as the field progresses. 

\section*{Acknowledgements}
AC thanks the ELLIS PhD program and the Danish Pioneer Centre for AI, DNRF grant number P1, for support.
ZA is supported by the ERC (853489-DEXIM).
OW is funded in part by the Novo Nordisk Foundation through the Center for Basic Machine Learning Research in Life Science (NNF20OC0062606).

\bibliographystyle{splncs04}
\bibliography{main}

\appendix

\title{Supplemental material} 
\titlerunning{Supplemental material: Geometry Fidelity for Spherical Images}
\author{}
\authorrunning{A.~Christensen et al.}
\institute{}

\maketitle

\section{Qualitative evaluation of \ours}
To qualitatively evaluate our proposed \oursFull, we compute FID and \ours\ on generated images from three different checkpoints of a finetuned text-to-image generative model. The model is based on a version of Imagen \cite{saharia2022photorealistic} trained on internal datasources, and finetuned using Dreambooth \cite{ruiz2023dreambooth} with a batch size of 16. We finetune the model on the 360-Indoor equirectangular image dataset \cite{Chou2019360IndoorTL} and use captions generated by a multimodal language model. This gives us 3252 image-caption pairs after removing duplicate and empty captions. 

The captions were generated by giving the multimodal model prompts with few-shot examples describing the content of corresponding equirectangular images, followed by keywords of e.g. style, lighting, and indoor/outdoor. An example of such a given few-shot example caption is: "\texttt{living room with couches, TV, coffee tables and fireplace. french style decoration, daylight, indoor}". The final prompt is then "\texttt{a panoramic view of a} $<$\texttt{caption}$>$".

Below, we show example equirectangular image generations from model checkpoints after 5000, 10000, and 20000 steps. The visualized generations are generated from the same prompts across the different checkpoints, where the corresponding prompts were selected randomly. Results show that the FID score is near-constant across the checkpoints (33.96, 35.42, 34.95, respectively). Further, although the example generations from the 5000 step model demonstrate that the model has issues constructing realistic geometry, the FID score is lowest for this checkpoint. On the contrary, \ours\ decreases monotonically over the checkpoints as geometry fidelity improves (63.39, 60.38, 55.07, respectively).

\begin{figure}[t]
  \centering
  \begin{subfigure}{0.49\linewidth}
    \centering
    \includegraphics[height=3.0cm]{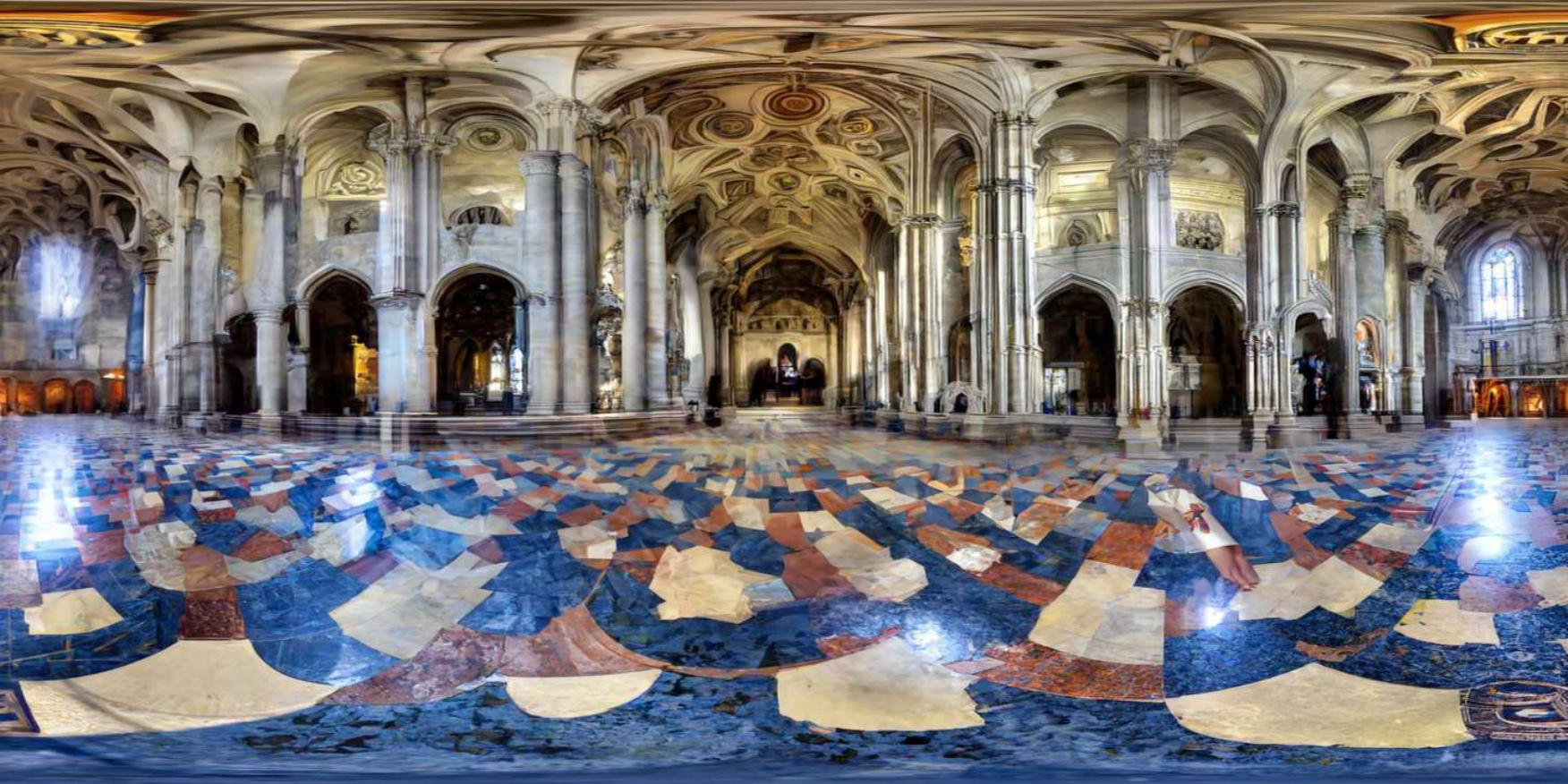}
    \label{fig:upscale-a}
  \end{subfigure}
  \begin{subfigure}{0.49\linewidth}
    \centering
    \includegraphics[height=3.0cm]{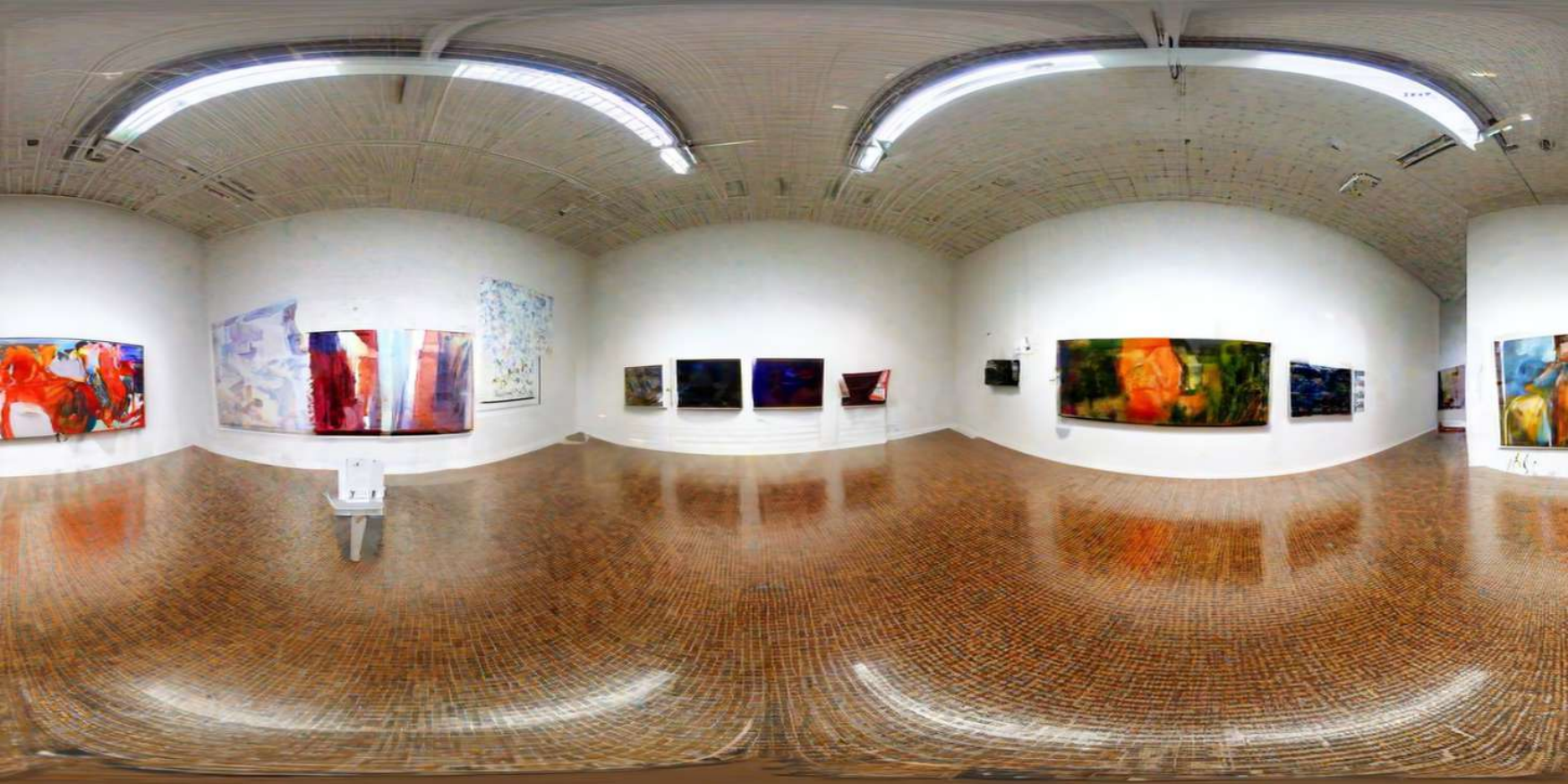}
    \label{fig:upscale-b}
  \end{subfigure}
  \begin{subfigure}{0.49\linewidth}
    \centering
    \includegraphics[height=1.7cm]{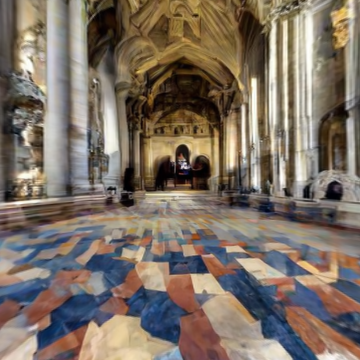}
    \includegraphics[height=1.7cm]{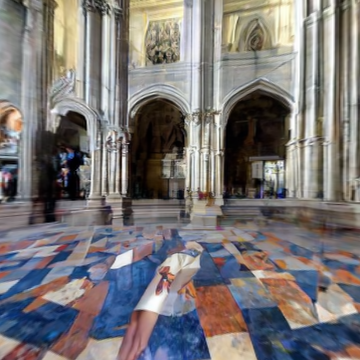}
    \includegraphics[height=1.7cm]{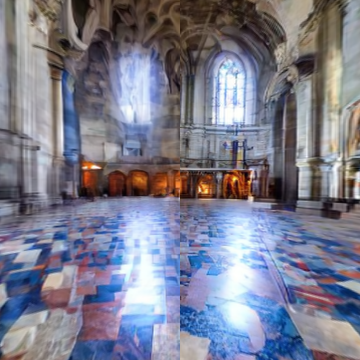}
    \includegraphics[height=1.7cm]{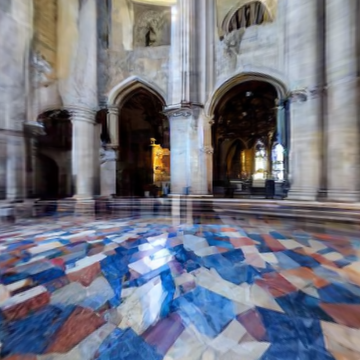}
    \includegraphics[height=1.7cm]{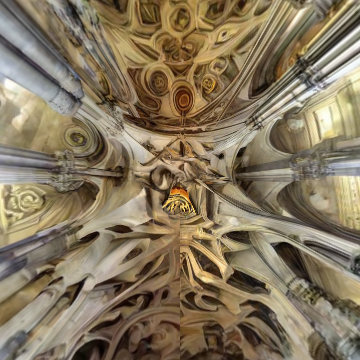}
    \includegraphics[height=1.7cm]{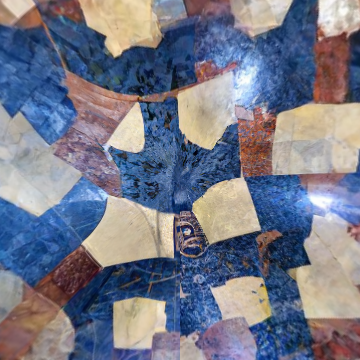}
    \label{fig:upscale-a}
    \vspace{1cm}
  \end{subfigure}
  \begin{subfigure}{0.49\linewidth}
    \centering
    \includegraphics[height=1.7cm]{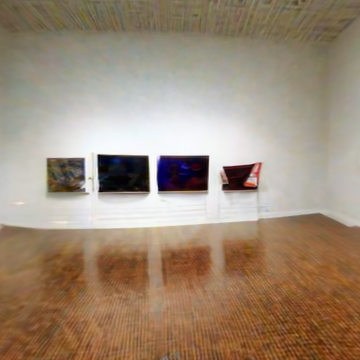}
    \includegraphics[height=1.7cm]{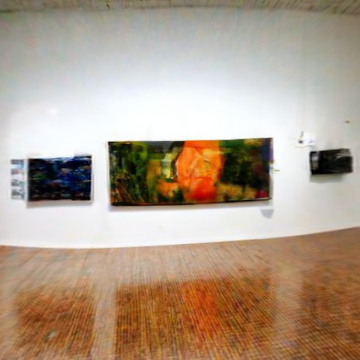}
    \includegraphics[height=1.7cm]{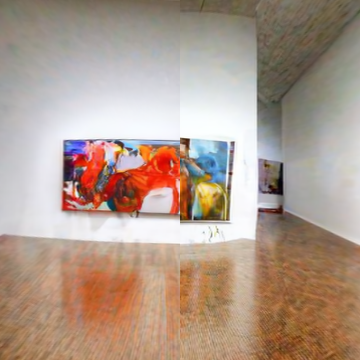}
    \includegraphics[height=1.7cm]{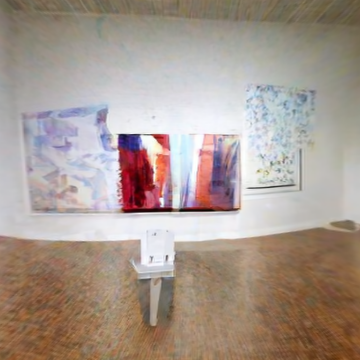}
    \includegraphics[height=1.7cm]{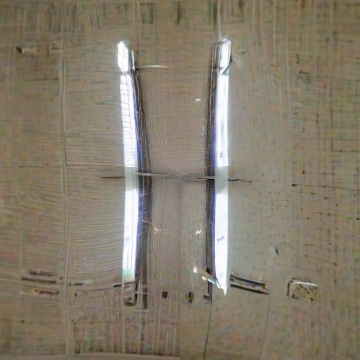}
    \includegraphics[height=1.7cm]{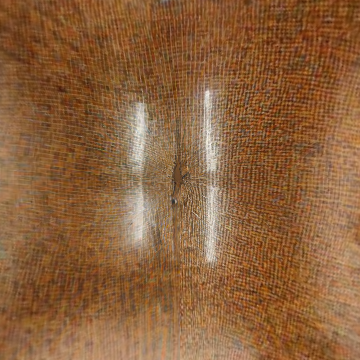}
    \label{fig:upscale-b}
    \vspace{1cm}
  \end{subfigure}
  \begin{subfigure}{0.49\linewidth}
    \centering
    \includegraphics[height=3.0cm]{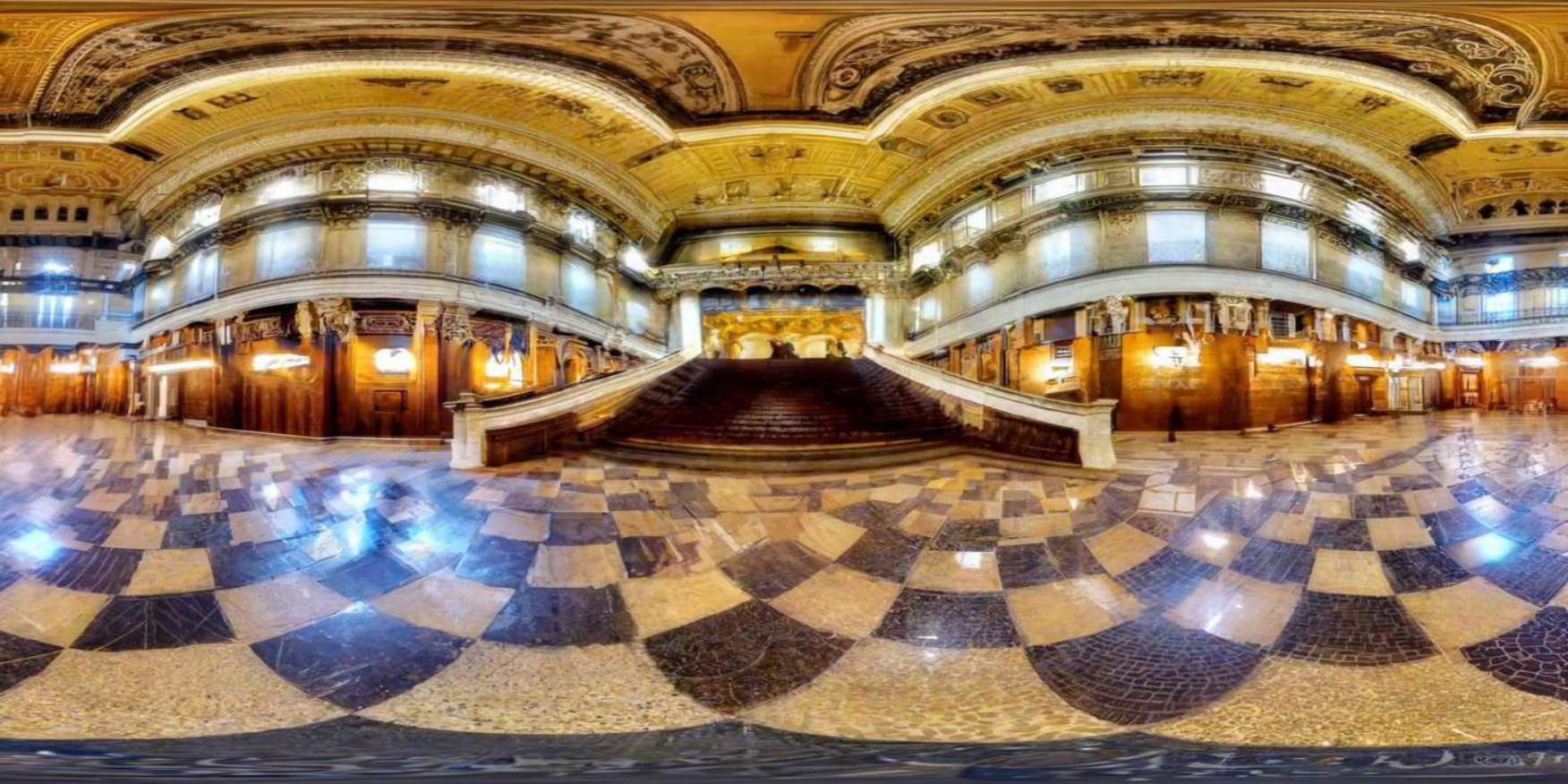}
    \label{fig:upscale-a}
  \end{subfigure}
  \begin{subfigure}{0.49\linewidth}
    \centering
    \includegraphics[height=3.0cm]{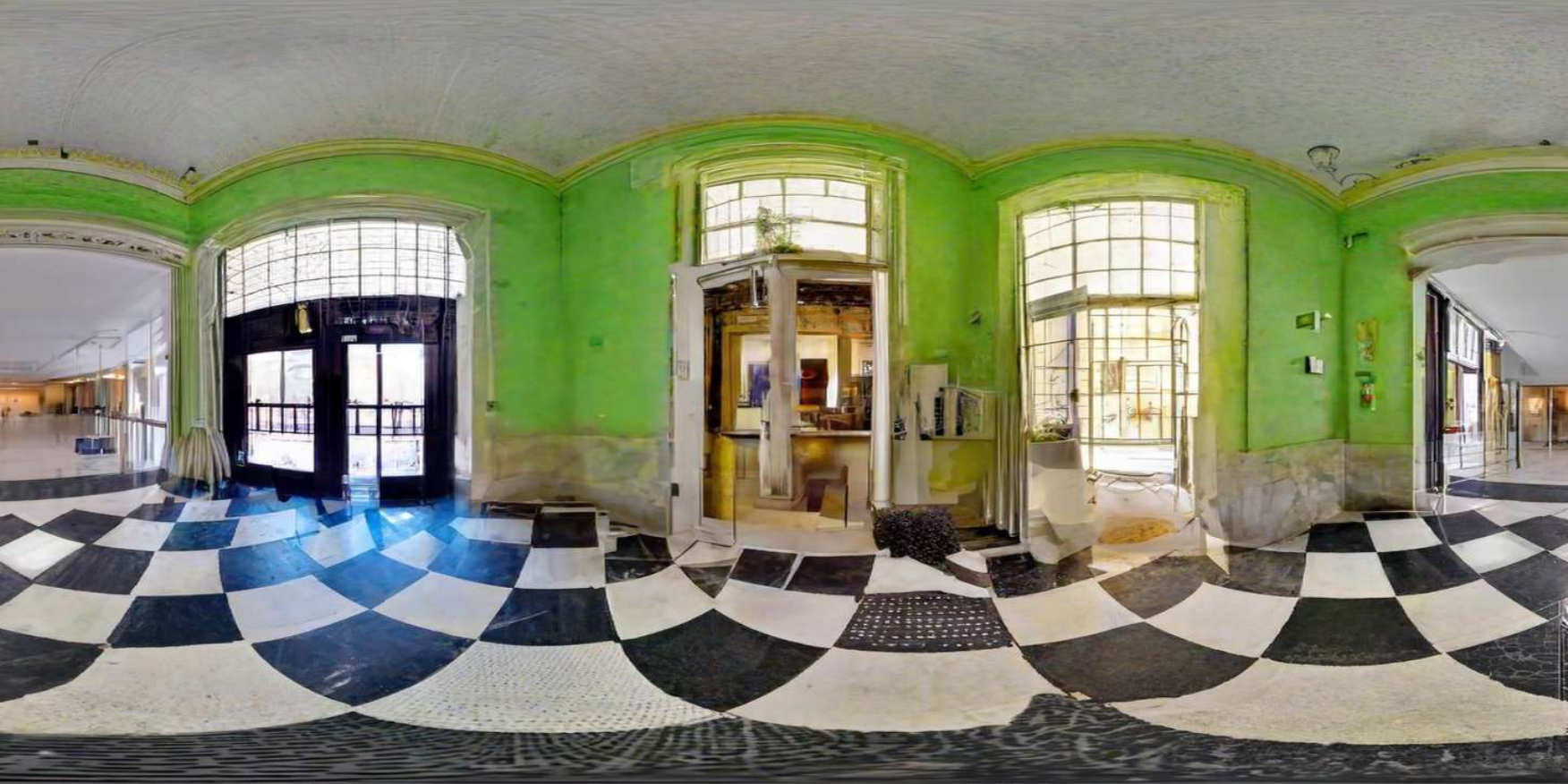}
    \label{fig:upscale-b}
  \end{subfigure}
  \noindent
  \begin{subfigure}{0.49\linewidth}
    \centering
    \noindent
    \includegraphics[height=1.7cm]{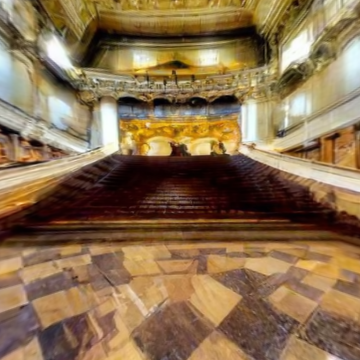}
    \includegraphics[height=1.7cm]{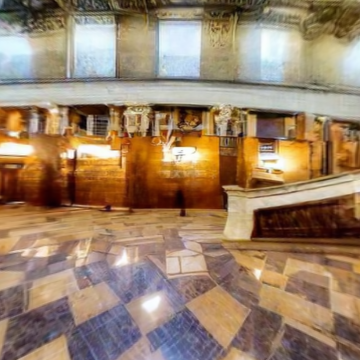}
    \includegraphics[height=1.7cm]{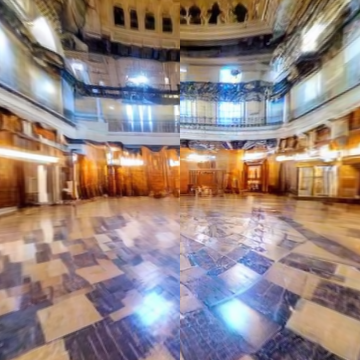}
    \includegraphics[height=1.7cm]{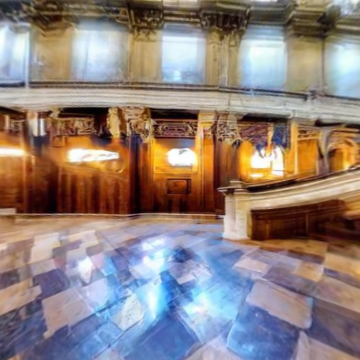}
    \includegraphics[height=1.7cm]{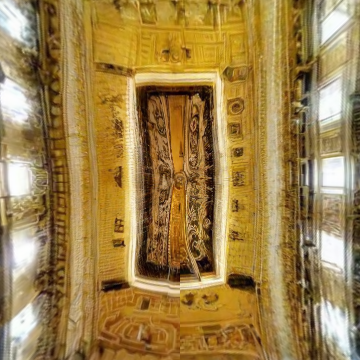}
    \includegraphics[height=1.7cm]{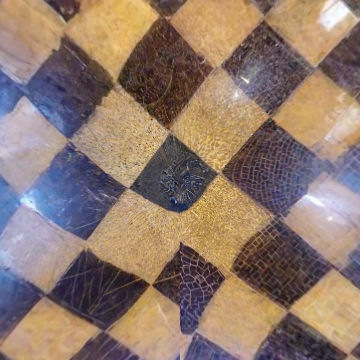}
    \label{fig:upscale-a}
  \end{subfigure}
\noindent
  \begin{subfigure}{0.49\linewidth}
    \centering
    \noindent
    \includegraphics[height=1.7cm]{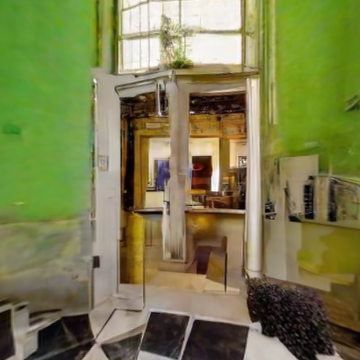}
    \includegraphics[height=1.7cm]{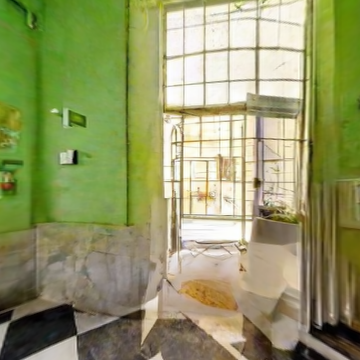}
    \includegraphics[height=1.7cm]{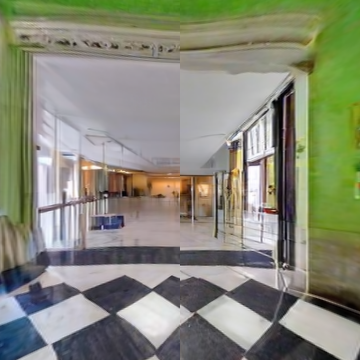}
    \includegraphics[height=1.7cm]{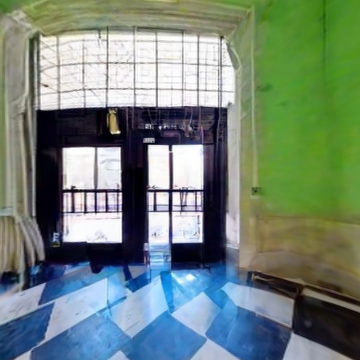}
    \includegraphics[height=1.7cm]{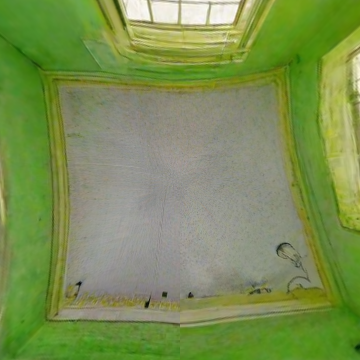}
    \includegraphics[height=1.7cm]{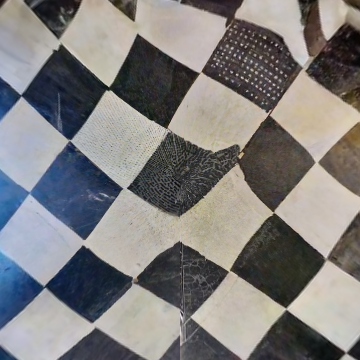}
    \label{fig:upscale-b}
  \end{subfigure}
  
  \caption{Four example equirectangular generations of a text-to-image model fine-tuned on 360-Indoor after 5000 steps. Under each generation we show the cubemap images to illustrate the geometry of the rendered views (top left to bottom right: front/right/back/left/up/down). FID is 33.96, \ours\ is 63.39.}
  \label{fig:supp:5000}. 
\end{figure}

\begin{figure}[t]
  \centering
  \begin{subfigure}{0.49\linewidth}
    \centering
    \includegraphics[height=3.0cm]{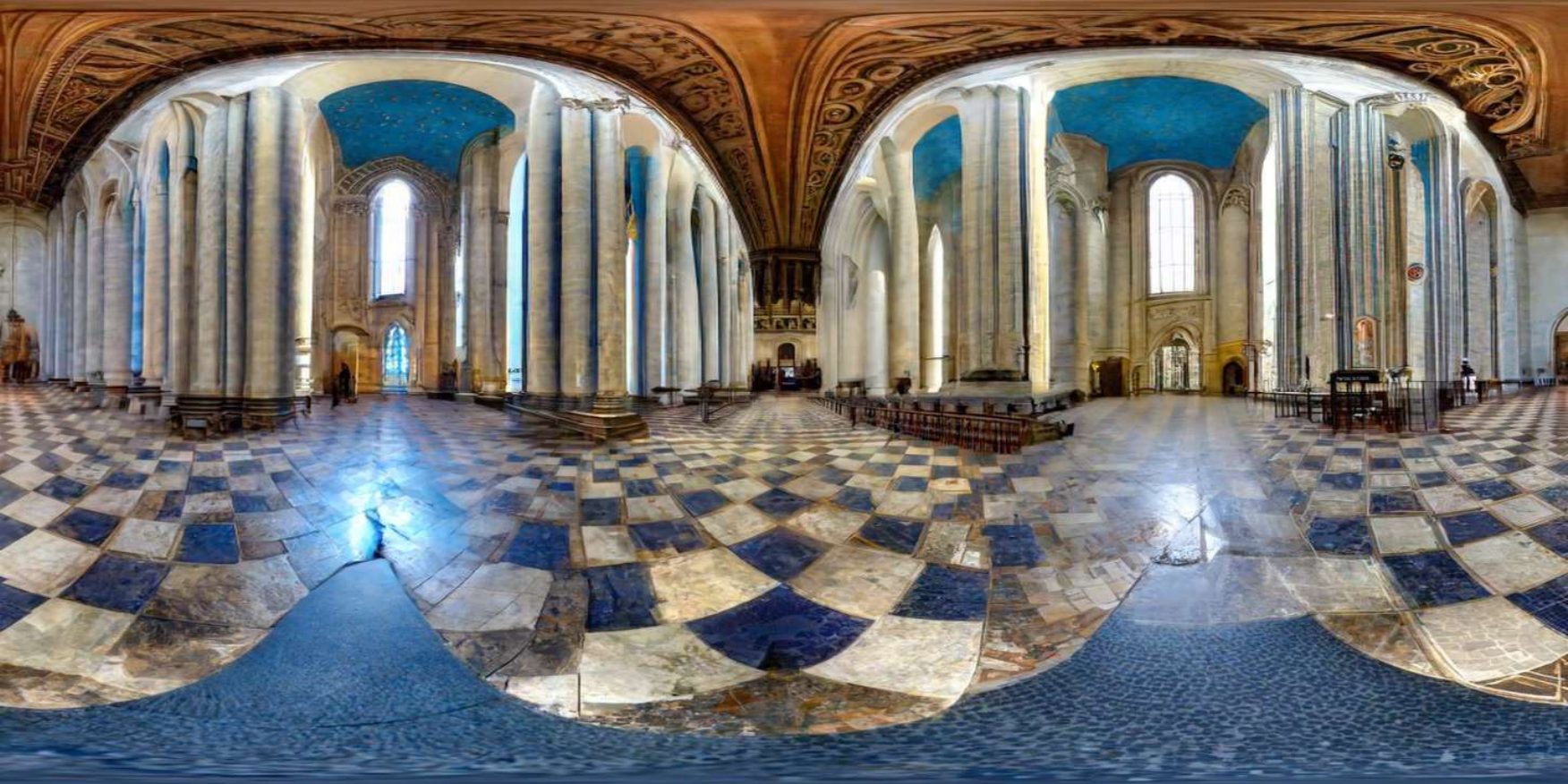}
    \label{fig:upscale-a}
  \end{subfigure}
  \begin{subfigure}{0.49\linewidth}
    \centering
    \includegraphics[height=3.0cm]{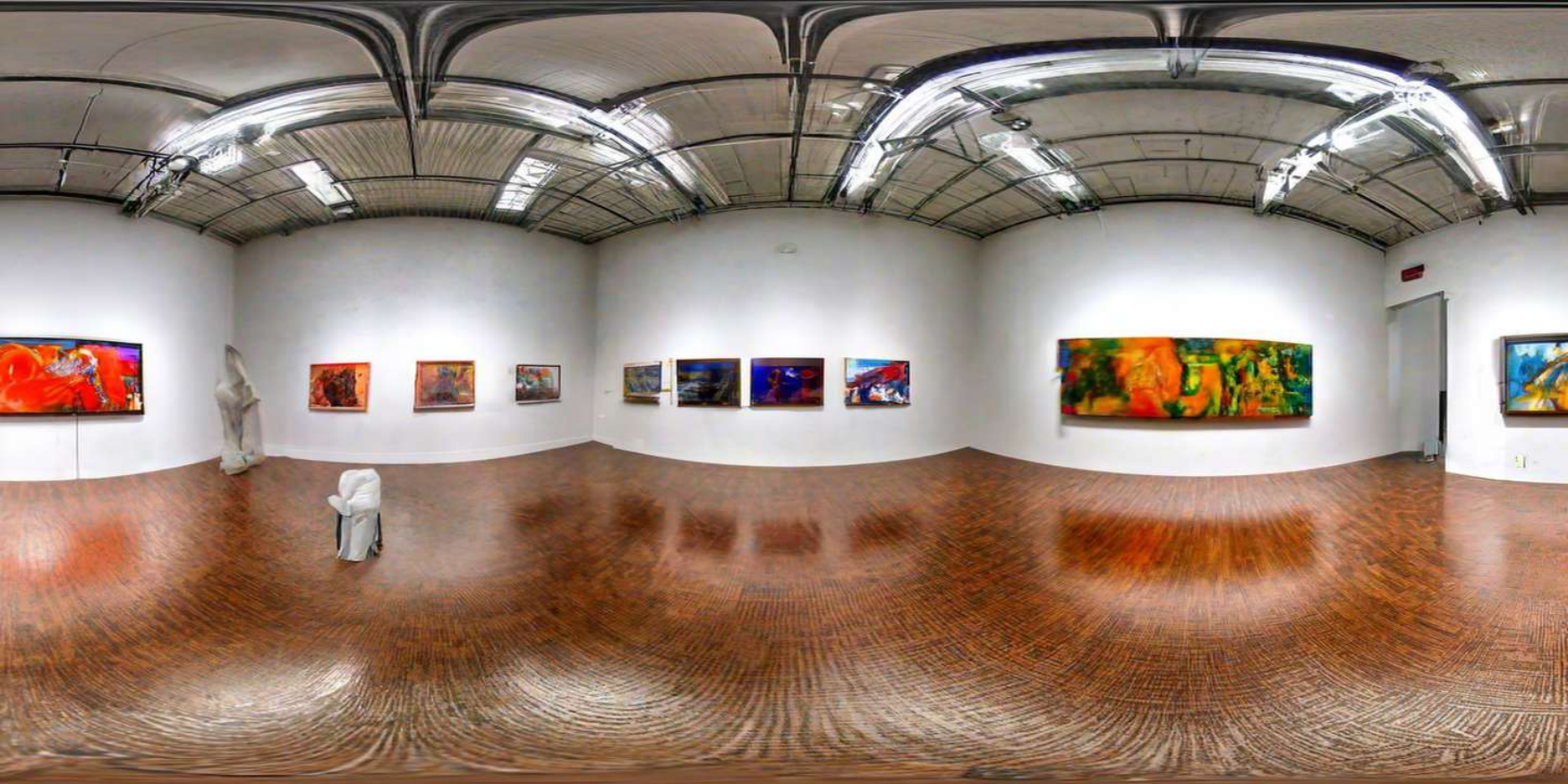}
    \label{fig:upscale-b}
  \end{subfigure}
  \begin{subfigure}{0.49\linewidth}
    \centering
    \includegraphics[height=1.7cm]{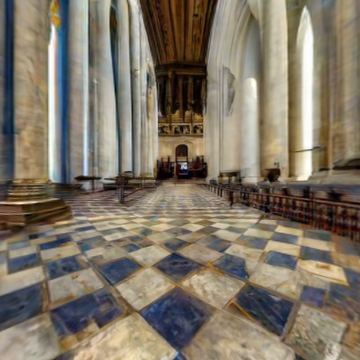}
    \includegraphics[height=1.7cm]{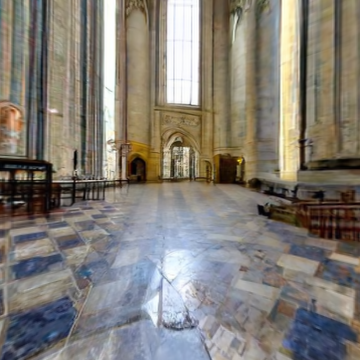}
    \includegraphics[height=1.7cm]{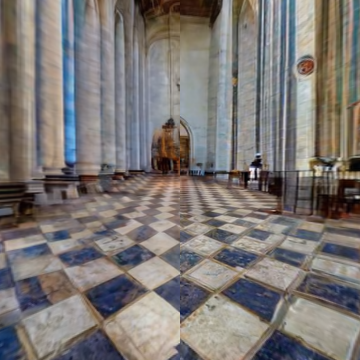}
    \includegraphics[height=1.7cm]{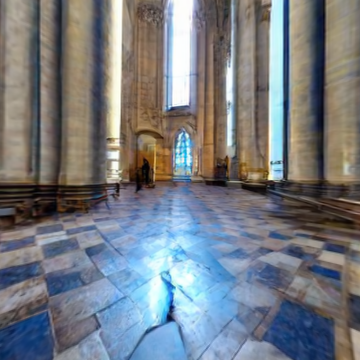}
    \includegraphics[height=1.7cm]{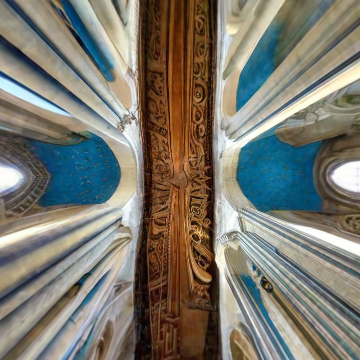}
    \includegraphics[height=1.7cm]{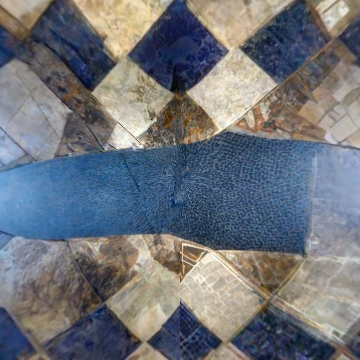}
    \label{fig:upscale-a}
    \vspace{1cm}
  \end{subfigure}
  \begin{subfigure}{0.49\linewidth}
    \centering
    \includegraphics[height=1.7cm]{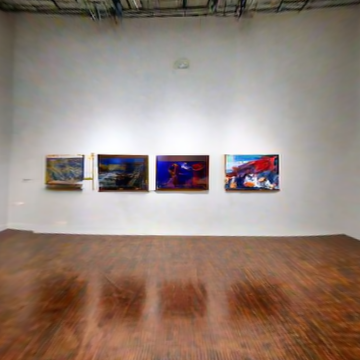}
    \includegraphics[height=1.7cm]{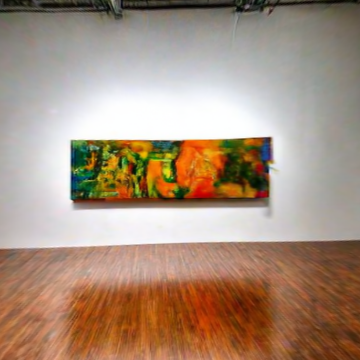}
    \includegraphics[height=1.7cm]{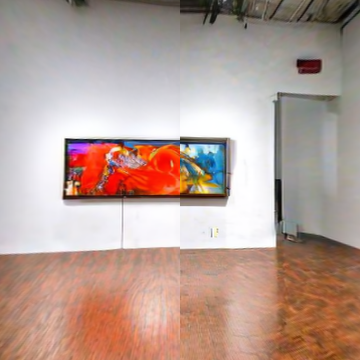}
    \includegraphics[height=1.7cm]{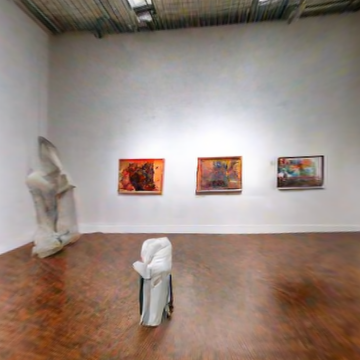}
    \includegraphics[height=1.7cm]{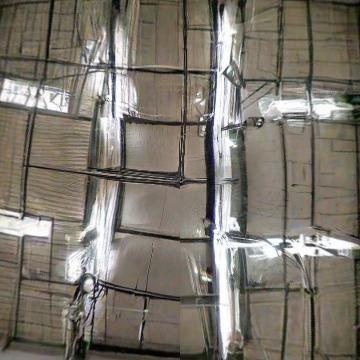}
    \includegraphics[height=1.7cm]{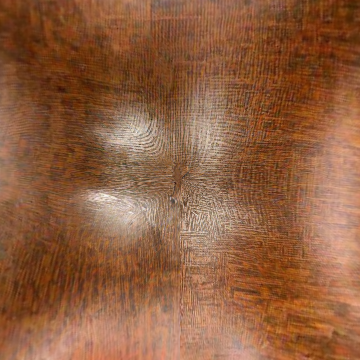}
    \label{fig:upscale-b}
    \vspace{1cm}
  \end{subfigure}
  \begin{subfigure}{0.49\linewidth}
    \centering
    \includegraphics[height=3.0cm]{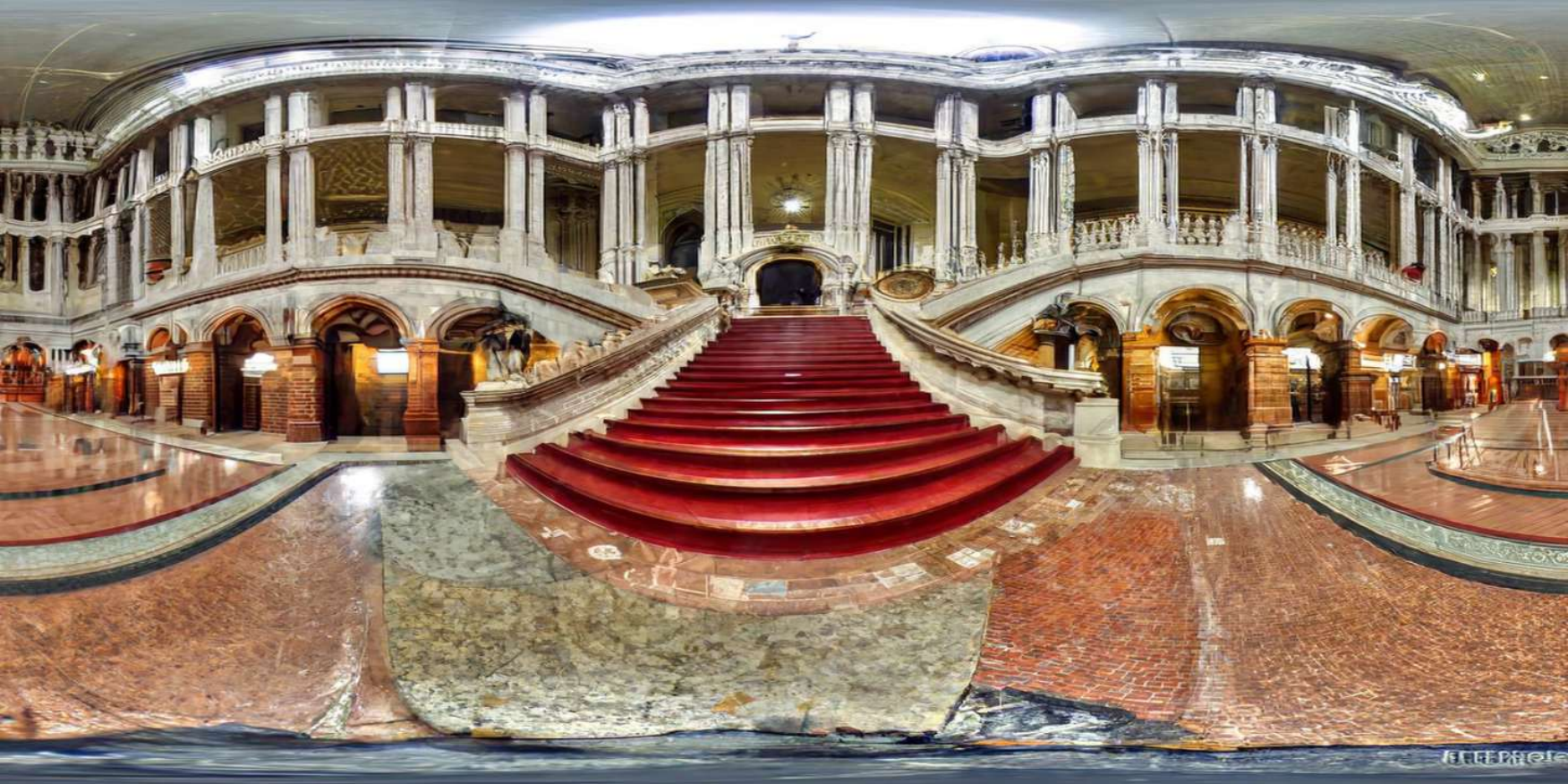}
    \label{fig:upscale-a}
  \end{subfigure}
  \begin{subfigure}{0.49\linewidth}
    \centering
    \includegraphics[height=3.0cm]{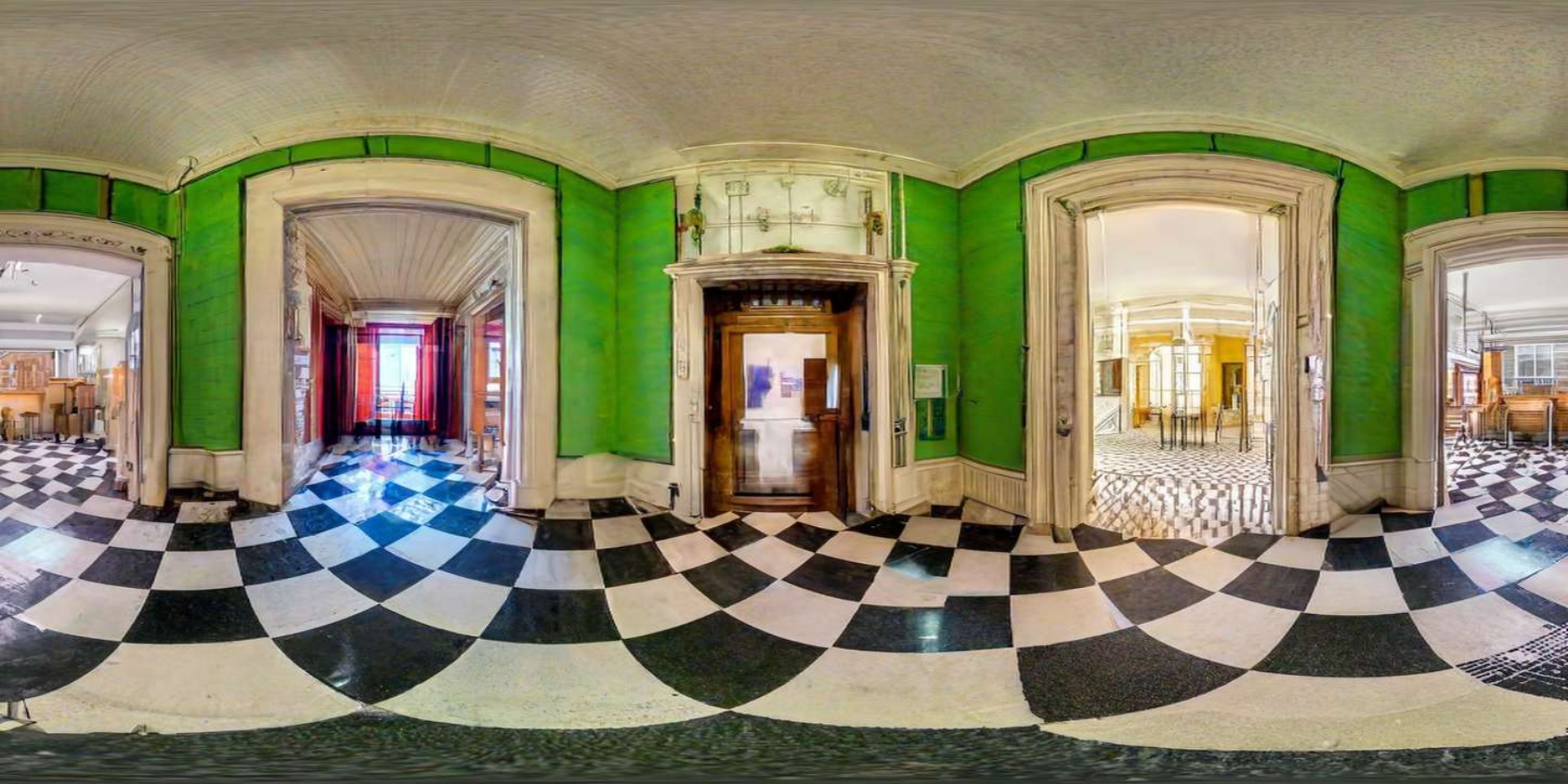}
    \label{fig:upscale-b}
  \end{subfigure}
  \noindent
  \begin{subfigure}{0.49\linewidth}
    \centering
    \noindent
    \includegraphics[height=1.7cm]{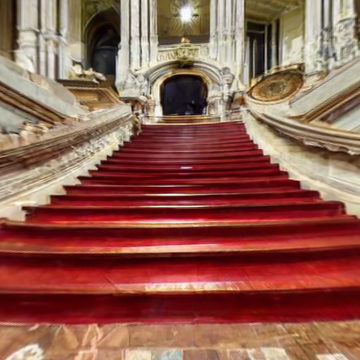}
    \includegraphics[height=1.7cm]{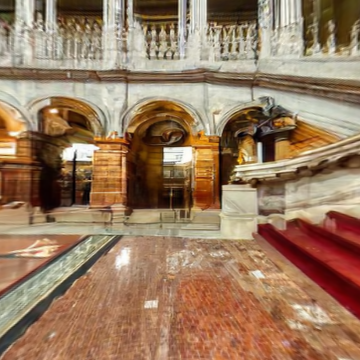}
    \includegraphics[height=1.7cm]{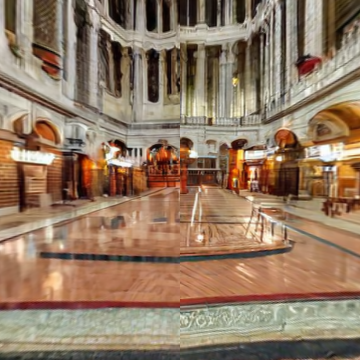}
    \includegraphics[height=1.7cm]{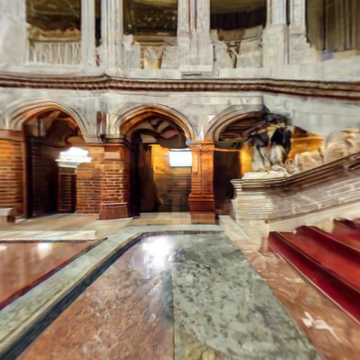}
    \includegraphics[height=1.7cm]{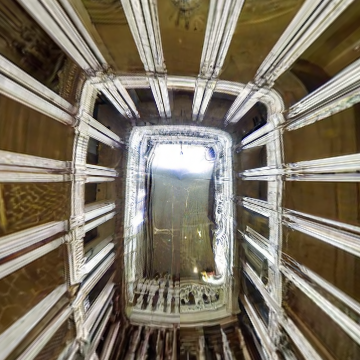}
    \includegraphics[height=1.7cm]{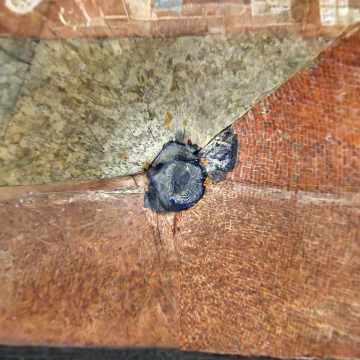}
    \label{fig:upscale-a}
  \end{subfigure}
\noindent
  \begin{subfigure}{0.49\linewidth}
    \centering
    \noindent
    \includegraphics[height=1.7cm]{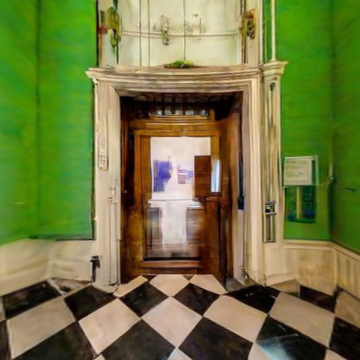}
    \includegraphics[height=1.7cm]{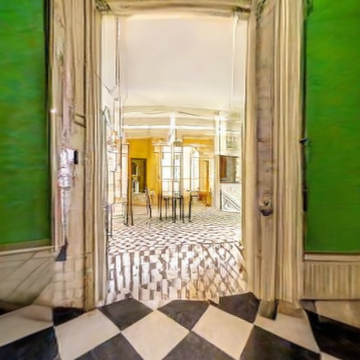}
    \includegraphics[height=1.7cm]{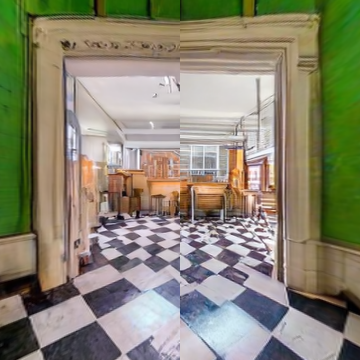}
    \includegraphics[height=1.7cm]{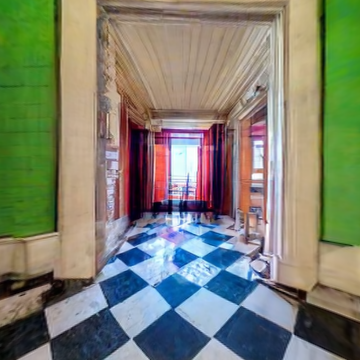}
    \includegraphics[height=1.7cm]{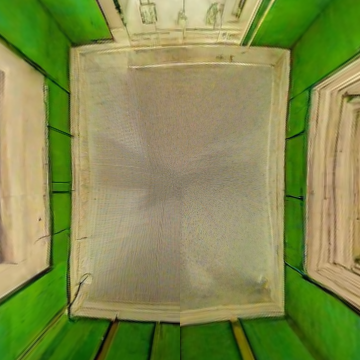}
    \includegraphics[height=1.7cm]{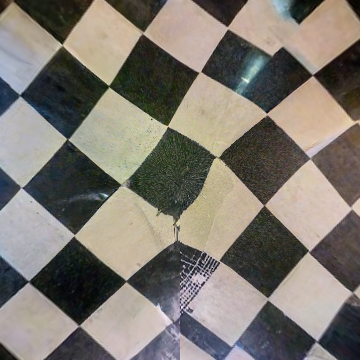}
    \label{fig:upscale-b}
  \end{subfigure}
  
  \caption{Four example equirectangular generations of a text-to-image model fine-tuned on 360-Indoor after 10000 steps. Under each generation we show the cubemap images to illustrate the geometry of the rendered views (top left to bottom right: front/right/back/left/up/down). FID is 35.42, \ours\ is 60.38.}
  \label{fig:supp:10000}. 
\end{figure}

\begin{figure}[t]
  \centering
  \begin{subfigure}{0.49\linewidth}
    \centering
    \includegraphics[height=3.0cm]{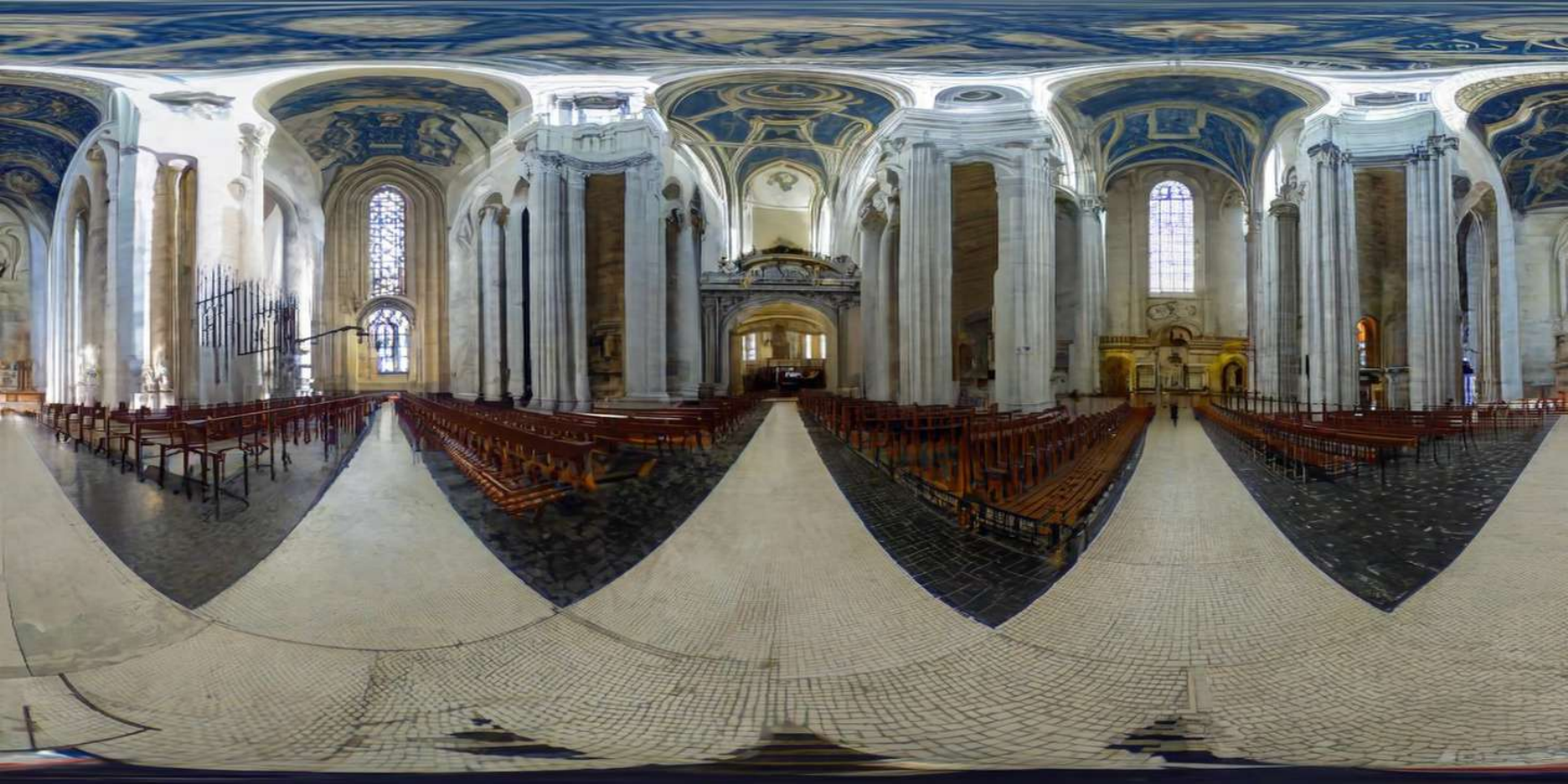}
    \label{fig:upscale-a}
  \end{subfigure}
  \begin{subfigure}{0.49\linewidth}
    \centering
    \includegraphics[height=3.0cm]{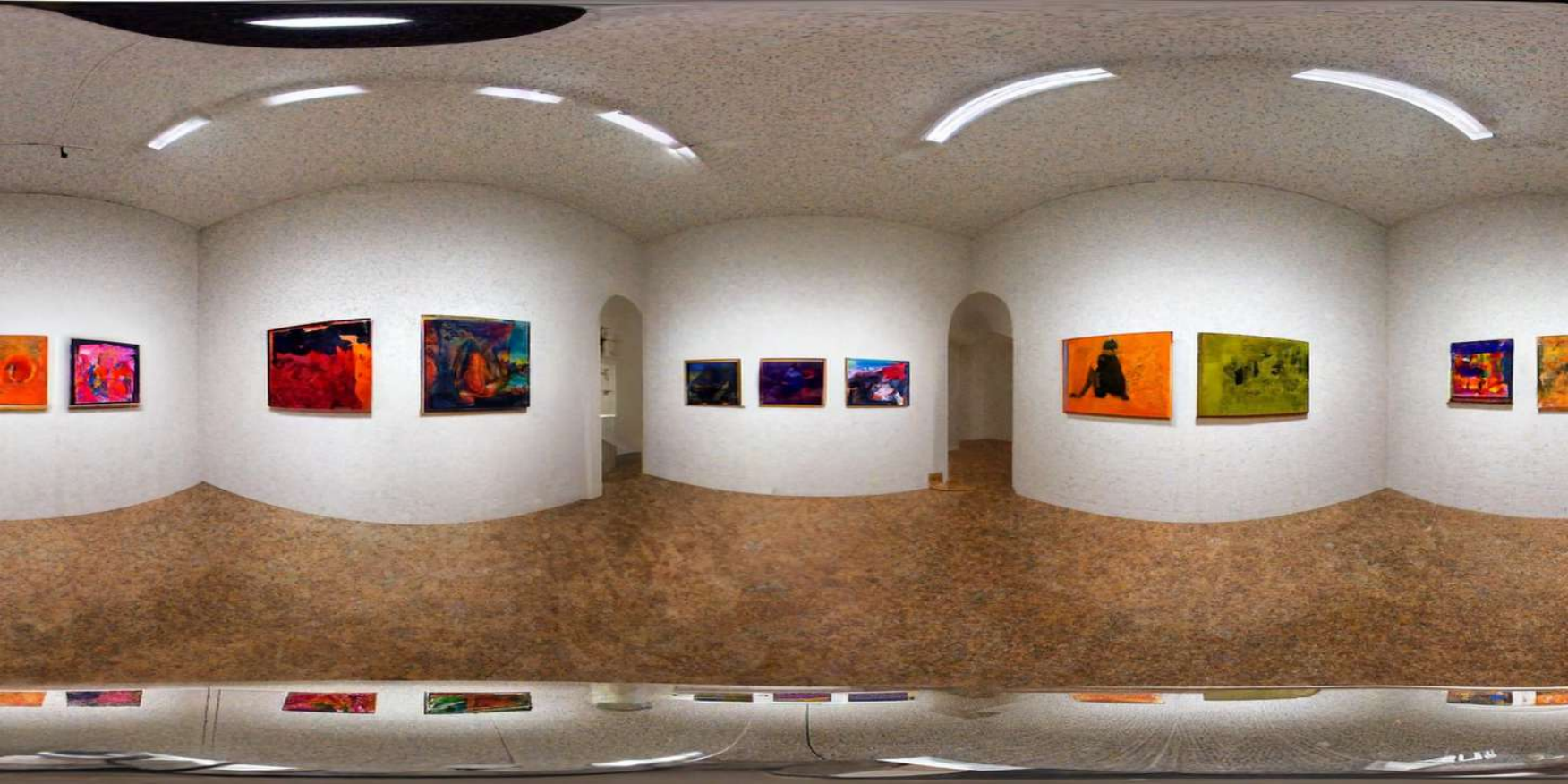}
    \label{fig:upscale-b}
  \end{subfigure}
  \begin{subfigure}{0.49\linewidth}
    \centering
    \includegraphics[height=1.7cm]{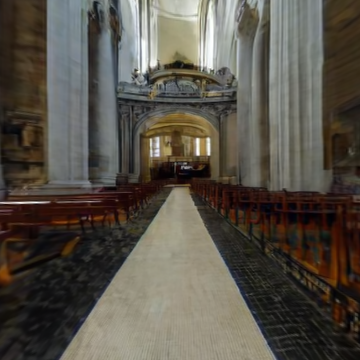}
    \includegraphics[height=1.7cm]{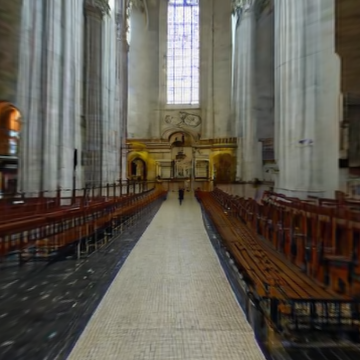}
    \includegraphics[height=1.7cm]{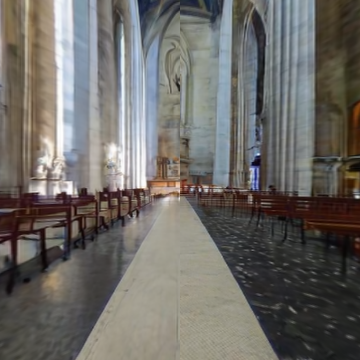}
    \includegraphics[height=1.7cm]{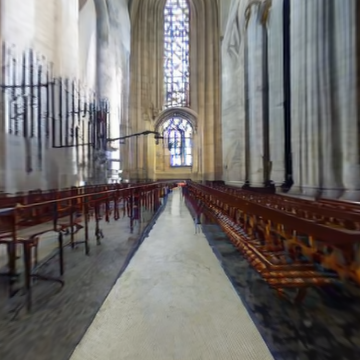}
    \includegraphics[height=1.7cm]{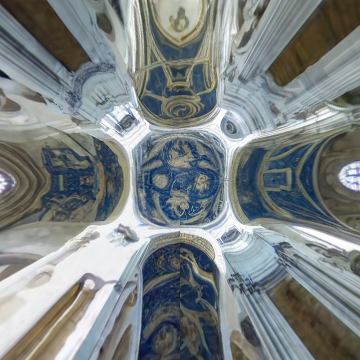}
    \includegraphics[height=1.7cm]{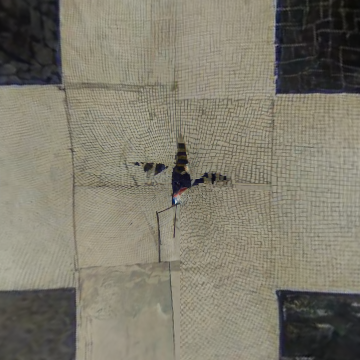}
    \label{fig:upscale-a}
    \vspace{1cm}
  \end{subfigure}
  \begin{subfigure}{0.49\linewidth}
    \centering
    \includegraphics[height=1.7cm]{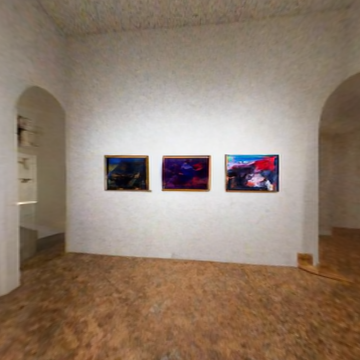}
    \includegraphics[height=1.7cm]{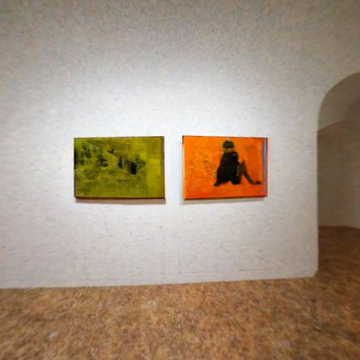}
    \includegraphics[height=1.7cm]{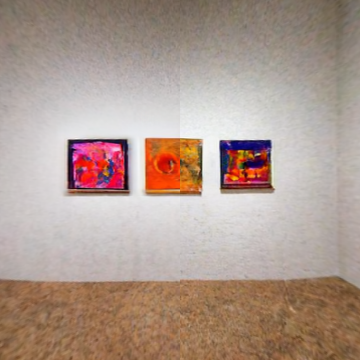}
    \includegraphics[height=1.7cm]{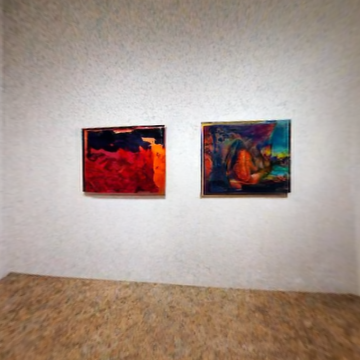}
    \includegraphics[height=1.7cm]{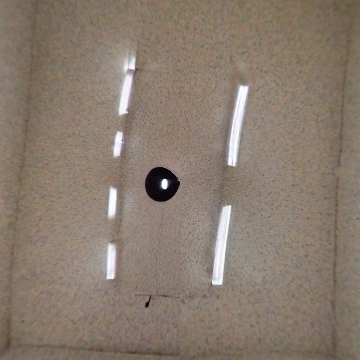}
    \includegraphics[height=1.7cm]{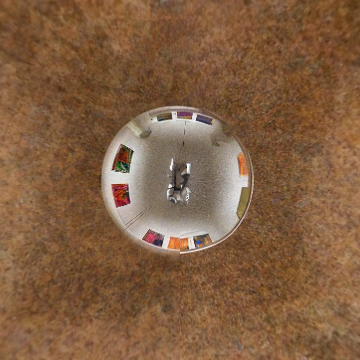}
    \label{fig:upscale-b}
    \vspace{1cm}
  \end{subfigure}
  \begin{subfigure}{0.49\linewidth}
    \centering
    \includegraphics[height=3.0cm]{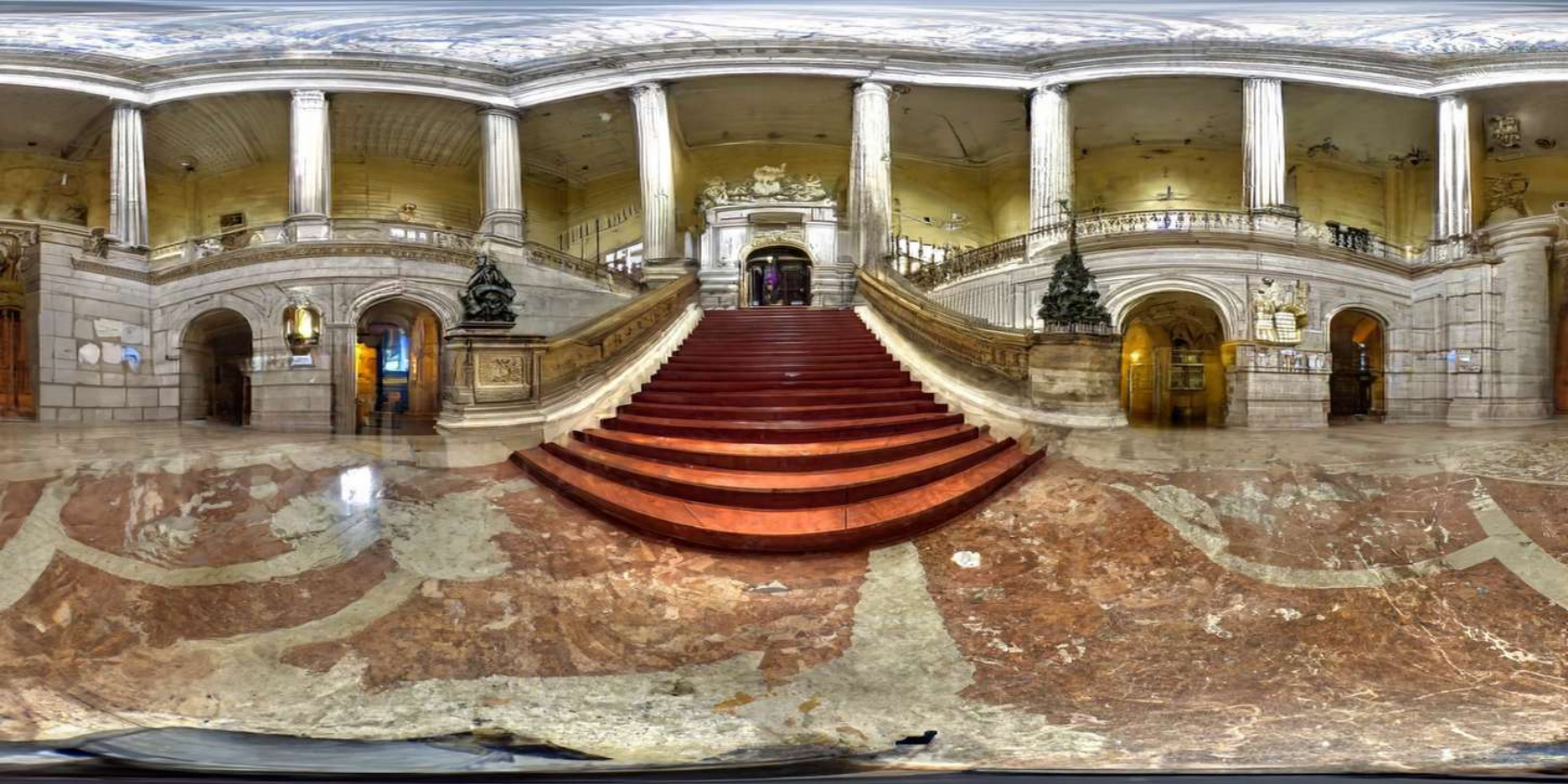}
    \label{fig:upscale-a}
  \end{subfigure}
  \begin{subfigure}{0.49\linewidth}
    \centering
    \includegraphics[height=3.0cm]{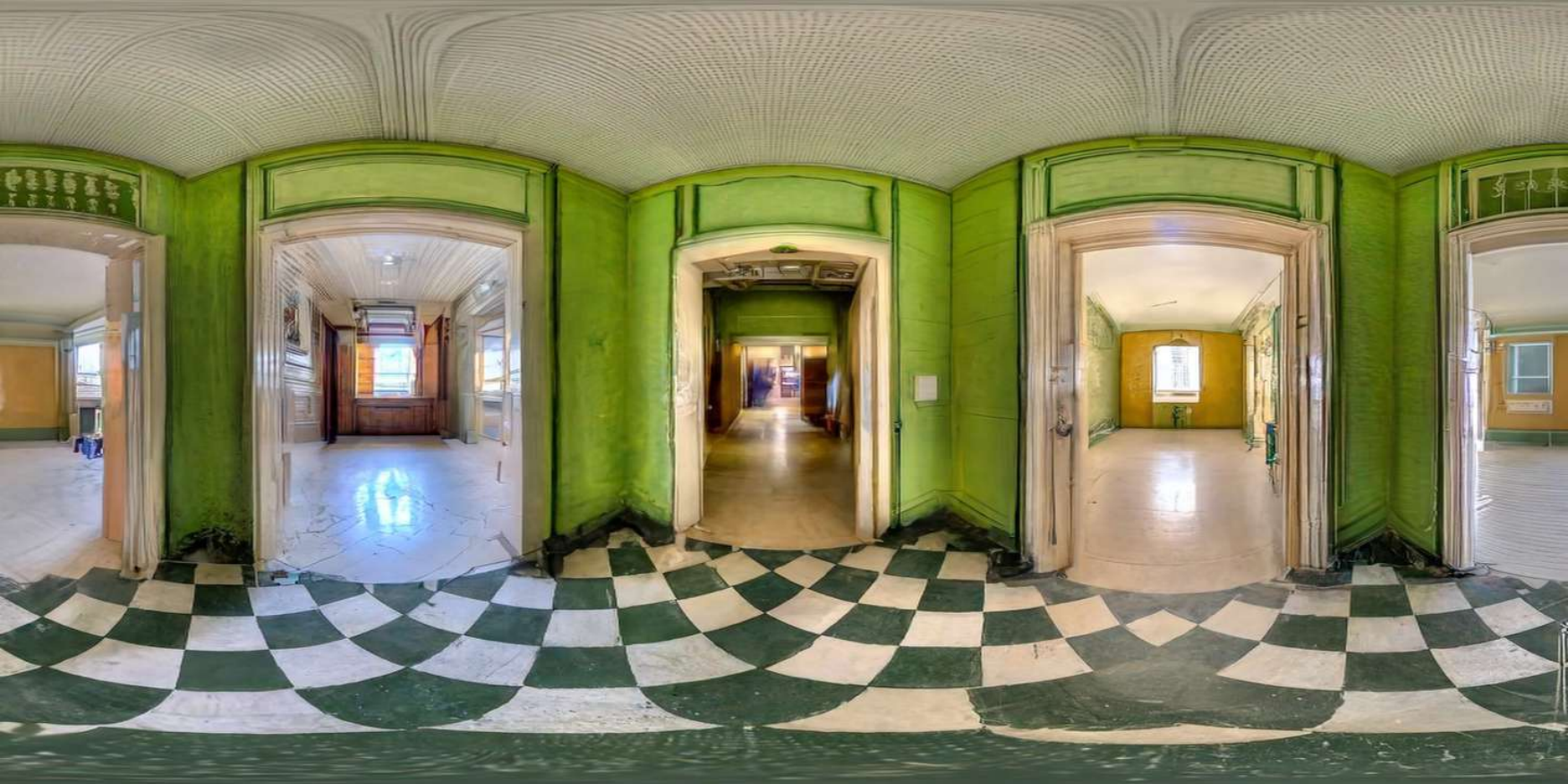}
    \label{fig:upscale-b}
  \end{subfigure}
  \noindent
  \begin{subfigure}{0.49\linewidth}
    \centering
    \noindent
    \includegraphics[height=1.7cm]{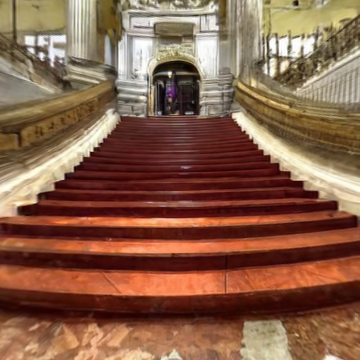}
    \includegraphics[height=1.7cm]{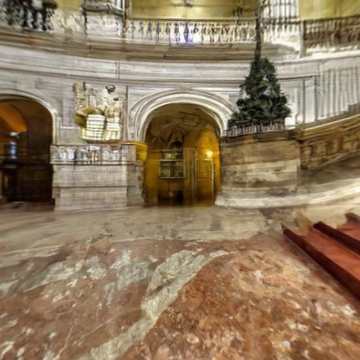}
    \includegraphics[height=1.7cm]{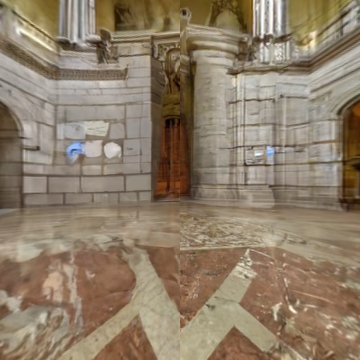}
    \includegraphics[height=1.7cm]{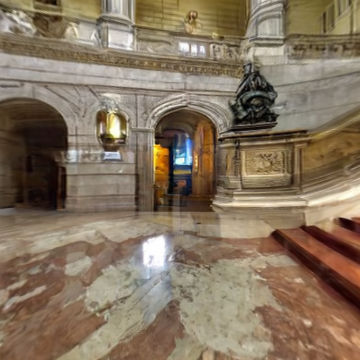}
    \includegraphics[height=1.7cm]{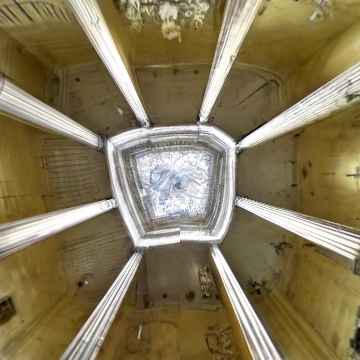}
    \includegraphics[height=1.7cm]{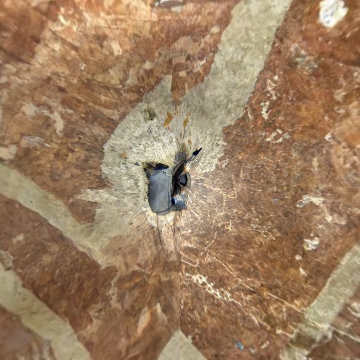}
    \label{fig:upscale-a}
  \end{subfigure}
\noindent
  \begin{subfigure}{0.49\linewidth}
    \centering
    \noindent
    \includegraphics[height=1.7cm]{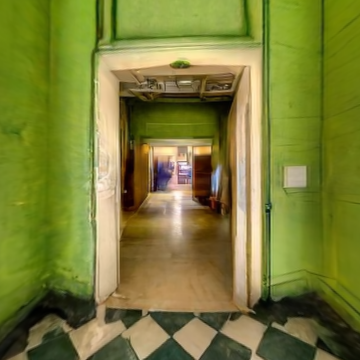}
    \includegraphics[height=1.7cm]{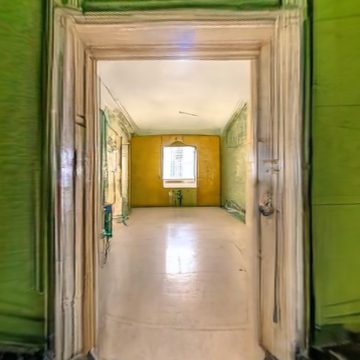}
    \includegraphics[height=1.7cm]{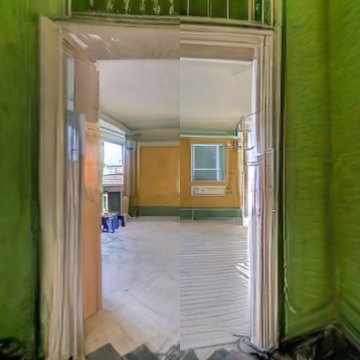}
    \includegraphics[height=1.7cm]{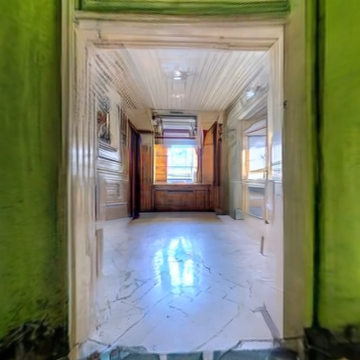}
    \includegraphics[height=1.7cm]{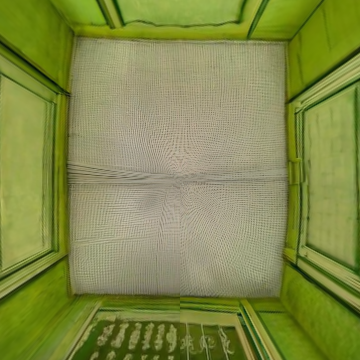}
    \includegraphics[height=1.7cm]{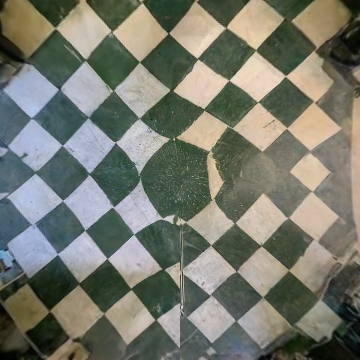}
    \label{fig:upscale-b}
  \end{subfigure}
  
  \caption{Four example equirectangular generations of a text-to-image model fine-tuned on 360-Indoor after 20000 steps. Under each generation we show the cubemap images to illustrate the geometry of the rendered views (top left to bottom right: front/right/back/left/up/down). FID is 34.95, \ours\ is 55.07.}
  \label{fig:supp:20000}. 
\end{figure}

\end{document}